\documentclass[preprint,12pt]{elsarticle}

\usepackage{graphicx}
\graphicspath{{./}}
\DeclareGraphicsExtensions{.pdf,.png,.jpg}

\usepackage{amsthm}
\usepackage{amssymb}
\usepackage{amsfonts}
\usepackage{amsmath}
\usepackage{nicefrac}
\usepackage{faktor}
\usepackage{url}

\usepackage{tabularx}

\journal{Pattern Recognition}

\usepackage{comment}
\usepackage{todonotes}
\usepackage[ruled,linesnumbered]{algorithm2e}
\newlength\mylen

\usepackage{multirow}
\usepackage{subcaption}
\usepackage{arydshln}
\usepackage[table,xcdraw]{}

\usepackage[normalem]{ulem}
\useunder{\uline}{\ul}{}

\newcommand{\etal}{\textit{et al.}}

\newcolumntype{Y}{>{\centering\arraybackslash}X}
\usepackage{acronym}
\acrodef{OMR}{Optical Music Recognition}
\acrodef{OCR}{Optical Character Recognition}
\acrodef{CRNN}{Convolutional Recurrent Neural Network}
\acrodef{ML}{Machine Learning}
\acrodef{DL}{Deep Learning}
\acrodef{CNN}{Convolution Neural Networks}
\acrodef{DNN}{Deep Neural Network}
\acrodef{SAE}{Selectional Auto-Encoder}
\acrodef{$F_{1}$}{F-measure}
\acrodef{RLE}{Run-length encoding}
\acrodef{TP}{True Positive}
\acrodef{FN}{False Negative}
\acrodef{FP}{False Positive}
\acrodef{5-CV}{5-fold cross-validation technique}
\acrodef{MAE}{Mean Absolute Error}
\acrodef{ReLU}{Rectified Linear Unit}
\acrodef{GRL}{Gradient Reversal Layer}
\acrodef{DA}{Domain Adaptation}
\acrodef{SVM}{Support Vector Machines}
\acrodef{LSTM}{Long Short-Term Memory}
\acrodef{FCNN}{Fully Convolutional Neural Network}
\acrodef{GAN}{Generative Adversarial Network}
\acrodef{DANN}{Domain-Adversarial Neural Network}
\acrodef{ReLU}{Rectifier Linear Unit}
\acrodef{IS}{Inception Score}

\definecolor{OliveGreen}{rgb}{0,0.38,0}
\definecolor{RedViolet}{rgb}{0.73, 0.2, 0.52}

\newcommand{\RevNote}[1]{\textcolor{black}{#1}}

\usepackage{pifont}

\newcommand{\secref}[1]{\mbox{Section~\ref{#1}}}
\newcommand{\tabref}[1]{\mbox{Table~\ref{#1}}}
\newcommand{\figref}[1]{\mbox{Figure~\ref{#1}}}

\newcommand{\algref}[1]{\mbox{Algorithm~\ref{#1}}}
\newcommand{\linealgref}[1]{\mbox{line~\ref{#1}}}

\newcommand{\SourceDS}{\text{$\mathcal{S}$}}%
\newcommand{\TargetDS}{\text{$\mathcal{T}$}}%
\newcommand{\XSource}{\text{$\mathcal{X}_{\text{S}}$}}%
\newcommand{\YSource}{\text{$\mathcal{Y}_{\text{S}}$}}%
\newcommand{\XTarget}{\text{$\mathcal{X}_{\text{T}}$}}%
\newcommand{\ProbSource}{\text{$\mathcal{P}_{\text{S}}$}}%
\newcommand{\ProbTarget}{\text{$\mathcal{P}_{\text{T}}$}}%
\newcommand{\HistTarget}{\text{$\mathcal{H}_{\text{T}}$}}%
\newcommand{\HistSource}{\text{$\mathcal{H}_{\text{S}}$}}%
\newcommand{\HistPrec}{\text{$\mathcal{H}_{\text{prec}}$}}%
\newcommand{\RHOth}{\text{$\rho_{\text{th}}$}}%
\newcommand{\RHOst}{\text{$\rho_{\text{S,T}}$}}%
\newcommand{\RHOss}{\text{$\rho_{\text{S,S}}$}}%

\newcommand{\ThresholdS}{\text{$\text{th}_\text{s}$}}%

\newcommand{\Precision}{$\text{P}$}%
\newcommand{\Recall}{$\text{R}$}%

\newcommand{\BinDANN}{\text{Bin-DANN}}%
\newcommand{\AutoBinDANN}{\text{AutoBin-DANN}}%

\newcommand{\Salzinnes}{\textsc{Salzinnes}}%
\newcommand{\Einsiedeln}{\textsc{Einsiedeln}}%
\newcommand{\Dibco}{\textsc{Dibco}}%
\newcommand{\PhiD}{\textsc{Phi}}%
\newcommand{\Palm}{\textsc{Palm}}%

\newcommand{\StoT}{~$\to$~}%

\SetArgSty{textnormal}

\begin{document}

\begin{frontmatter}


\title{Unsupervised Neural Domain Adaptation for Document Image Binarization}



\author[add1]{Francisco J. Castellanos\corref{cor1}}
\ead{fcastellanos@dlsi.ua.es}
\cortext[cor1]{Corresponding author.}

\author[add1]{Antonio-Javier Gallego}
\ead{jgallego@dlsi.ua.es}

\author[add1]{Jorge Calvo-Zaragoza}
\ead{jcalvo@dlsi.ua.es}

\address[add1]{Department of Software and Computing Systems, University of Alicante, Carretera San Vicente del Raspeig s/n, 03690 Alicante, Spain}

\begin{abstract}
Binarization is a well-known image processing task, whose objective is to separate the foreground of an image from the background. One of the many tasks for which it is useful is that of preprocessing document images in order to identify relevant information, such as text or symbols. The wide variety of document types, alphabets, and formats makes binarization challenging. There are multiple proposals with which to solve this problem, from classical manually-adjusted methods, to more recent approaches based on machine learning. The latter techniques require a large amount of training data in order to obtain good results; however, labeling a portion of each existing collection of documents is not feasible in practice. This is a common problem in supervised learning, which can be addressed by using the so-called Domain Adaptation (DA) techniques. These techniques take advantage of the knowledge learned in one domain, for which labeled data are available, to apply it to other domains for which there are no labeled data. This paper proposes a method that combines neural networks and DA in order to carry out unsupervised document binarization. However, when both the source and target domains are very similar, this adaptation could be detrimental. Our methodology, therefore, first measures the similarity between domains in an innovative manner in order to determine whether or not it is appropriate to apply the adaptation process. The results reported in the experimentation, when evaluating up to 20 possible combinations among five different domains, show that our proposal successfully deals with the binarization of new document domains without the need for labeled data. 

\end{abstract}

\begin{keyword}
Binarization \sep Machine Learning \sep Domain Adaptation \sep Adversarial training
\end{keyword}

\end{frontmatter}


\section{Introduction}
\label{sec:introduction}

In the context of documents, digital transcription is the process of exporting the information that is physically present on pages into a format that can be processed by a computer. Its objective is to preserve, and often disseminate, the content of these documents~\cite{doermann2014handbook}. This process could be performed manually, but this would be a tedious and error-prone task. One solution is the development of systems that are capable of automatically extracting the content~\cite{textLineDetEnrique2012,bainbridge01challenge} and subsequently encoding it into a structured digital format~\citep{krallinger2005text,wang2011end,hankinson2012creating}. The wide variety of document types, such as old books, medical reports or even handwritten music scores, makes automating this transcription task very complex, since it is necessary for the system to identify the relevant information to be extracted from each type of document.

One of the most common steps in document image processing is binarization. This process reduces the image to a binary representation (that is, a black and white image) by segmenting the relevant content---such as text, ornaments, or other types of symbols, which will depend on the document---and separating it from the background and other artifacts, such as ink stains, shadow-through, bleed-through, or other types of degradation. This process is often a key aspect in document image analysis workflows \cite{LOULOUDIS20083758,HE20154036,GIOTIS2017310}. The relevant literature contains multiple heuristic methods based on establishing either a local or global threshold with which to perform the process~\cite{sauvola2000adaptive,he2019deepotsu}. 

Binarization can also be formulated as a supervised machine learning task \cite{kita2010binarization,pastor-pellicer15insights}, in which a model is trained to classify each pixel of the image. This strategy, in addition to providing competitive results \cite{CALVOZARAGOZA201937}, comprises a more generalizable approach that automatically adjusts the model to each new domain by learning from labeled data. This formulation, however, entails the need for labeled data, which are not always available. Moreover, the models learned with these data are highly dependent on them, so they perform well only for the same domain, or at best, for similar ones. This means that, for each new type of document, it is necessary to retrain the method by using new labeled data from that domain, resulting in an inefficient approach in practice.

The issue mentioned above is a common problem in supervised learning, and there is consequently an open line of research, referred to as \ac{DA}, that studies how to mitigate it~\cite{wang2018deep}.
\ac{DA} algorithms propose mechanisms with which to apply the knowledge learned from a domain, for which labeled data are available, to other unlabeled domains. It is, therefore, necessary only to have a labeled source domain and perform the adaptation process to the new target domains without having to label more data.

This paper proposes a neural network approach that uses unsupervised \ac{DA} in order to binarize documents. This approach specifically modifies the \ac{SAE} network \cite{CALVOZARAGOZA201937}, previously used for binarization, so as to integrate it into a \ac{GRL} \cite{ganin2016domain}, thus allowing it to learn domain-invariant features and binarize documents from new domains without using new labeled data. 

\RevNote{Furthermore, in document binarization, it is likely that a model trained for one domain works well in many other (target) domains. Although this assumption may occur in other contexts related to semantic segmentation, this is more common in binarization due to the presence of the two same categories in all cases: ink and background. Both categories depict some common characteristics across all types of documents. This makes the situation mentioned above, with a source model behaving well in a new domain, more likely to happen for document image binarization.}
We will demonstrate that a high similarity between the source and the target domains may be detrimental to the adaptive learning process. Given this, and depending on the features of the new domain, it may not be appropriate to carry out the DA process. An innovative mechanism with which to measure the similarity between binarization domains is also proposed in order to evaluate the new data before adapting the features learned by the network. 

Our proposal is evaluated with five datasets that contain different domains, such as old handwritten text documents, musical scores or Balinese palm leaf manuscripts. The results obtained are compared with those obtained by the learning-driven state-of-the-art architecture proposed in~\cite{CALVOZARAGOZA201937}. 

The present work is organized as follows: a literature review is provided in Section~\ref{sec:background}, while the proposed approach is described in Section~\ref{sec:methodology} and the details of the experimental setup and the analysis of results are shown in Section~\ref{sec:experimental_setup}. Finally, our conclusions and future work are discussed in Section~\ref{sec:conclusions}.

\section{Background}
\label{sec:background}

\subsection{Binarization}
\label{sec:bin_bg}

A binarization process traditionally uses a hand-set threshold to separate the relevant information regarding the image from the rest of the content. However, the pixel values of what is relevant may change from one type of document to another, signifying that the threshold also depends on the type of document and its features (color, lighting, and so on). This approach, therefore, lacks the flexibility required for it to be applied directly to all types of documents without having to be adjusted manually each time.

On the one hand, many heuristic methods that estimate an optimal threshold from the features of the image have been developed. There are, for example, methods that calculate a global threshold or that subdivide the image into regions and calculate local thresholds for each of them. One such method is that of Otsu ~\cite{otsu1979threshold}, which employs the histogram of the grayscale image in order to compute a global threshold; Perez and Gonzalez~\cite{perez1987iterative}, meanwhile, proposed a method based on the expansion of Taylor's series to deal with images with non-uniform illumination. It is also worth mentioning the recent adaptive version of Otsu's method proposed by Moghaddam and Cheriet~\cite{moghaddam2012adotsu}, which binarizes images using a grid-based strategy to estimate a background map. There are many other methods that calculate a local rather than a global threshold. This concept was first proposed by Niblack~\cite{niblack1985introduction}, and computes one threshold per pixel on the basis of its neighborhood. The most common extensions to this are that of Sauvola~\cite{sauvola2000adaptive} and that proposed by Wolf \etal~\cite{wolf02text}, both of which employ more complex equations in order to define the local threshold calculation. There is also a binarization method based on Markov random fields~\cite{mishra2011mrf}, or the method proposed by Jia \etal~\cite{jia2018degraded}, which is used to binarize degraded documents and which computes local thresholds on the basis of structural symmetric pixels. The reader is referred to the available surveys on document image binarization for further literature on heuristic methods \cite{7155946,sulaiman2019,TensmeyerM20}.

On the other hand, there are methods that formulate binarization as a supervised learning problem. For example, Kita and Wakahara~\cite{kita2010binarization} proposed an approach that binarizes images using a combination of $k$-means clustering and \ac{SVM}~\cite{hearst1998support}; Chou \etal~\cite{chou2010binarization} binarize photographed documents by subdividing the image into different regions and calculating local thresholds by using \ac{SVM}, and Xiong \etal~\cite{xiong2018degraded} applied a similar method in order to binarize degraded scanned documents. Of the solutions based on machine learning, those that use deep neural networks~\cite{lecun2015deep} are especially relevant owing to their good performance. For instance, \ac{CNN} have been analyzed in order to classify each pixel of an image into a binary value~\cite{pastor-pellicer15insights}; Afzal \etal~\cite{afzal2015document} proposed a \ac{LSTM}~\cite{greff2016lstm} based method that achieves excellent results, even with degraded documents; Milletari \etal~\cite{milletari2016v} proposed a new \ac{FCNN} model with which to segment medical images; the PDNet approach~\cite{ayyalasomayajula2019pdnet} formulates binarization as an energy minimization problem, in which an unrolled primal-dual network based on \ac{FCNN} is trained to minimize the labeling cost for each pixel; 
He and Schomaker~\cite{he2019deepotsu} combined a heuristic method and a deep neural network for binarization: the neural network first processes the image in order to reduce the degradation level, after which Otsu's method is applied in order to binarize the document. Another work is that of Calvo and Gallego~\cite{CALVOZARAGOZA201937}, who proposed a method that uses a \ac{FCNN}-based \ac{SAE} architecture to efficiently binarize images of different collections, such as handwritten text documents or music manuscripts.

However, despite the good results reported for the solutions based on machine learning, and especially on deep learning, they all have the same major drawback: the need for sufficient annotated samples from each document collection. As stated above, this information is not always available, and manually labeling documents can be a costly task, which is not a scalable solution in practice. The development of techniques that allow these models to be used in new domains without requiring labeled data is, therefore, of particular interest.

\subsection{Domain Adaptation}
\label{sec:da_bg}

\ac{DA} techniques attempt to use the knowledge learned from a domain for which labeled data are available---referred to as the \textit{source domain}---in a different (but related) domain for which labeled data are not available---referred to as the \textit{target domain}. This can be achieved by using several strategies, which can be grouped into three main categories~\cite{wang2018deep}:

\begin{itemize}
\item \textit{Divergence-based \ac{DA}}: A domain-invariant feature representation is obtained by minimizing a measure of divergence between two corpora~\cite{yan2017mind,sun2015return,shen2017wasserstein}. For example, Rozantsev \etal~\cite{rozantsev2018beyond} proposed a pair of parallel neural architectures that regularize the loss function by using a Maximum Mean Discrepancy metric~\cite{yan2017mind}. Another example is the work of Sun and Saenko~\cite{sun2016deep}, which uses the Correlation Alignment metric~\cite{sun2015return} to minimize correlations between source and target domains. The DeepJDOT approach by Damodaran \etal~\cite{bhushan2018deepjdot} learns both a classifier and a common representation of the source and target domains by using loss functions based on the Optimal Transport theory~\cite{villani2008optimal}. 

\item \textit{Reconstruction-based \ac{DA}}: The objective of this strategy is to obtain a common representation of the data in order to use the same classifier for both source and target domains. One example of this is the Deep Reconstruction Classification Network~\cite{ghifary2016deep}, which uses a multi-task learning approach to learn an intermediate feature representation so as to classify both domains. Another is the proposal of Isola \etal~\cite{isola2017image}, which employs a conditional \ac{GAN} to transform one domain into another using an encoder-decoder or a U-Net architecture.

\item \textit{Adversarial-based \ac{DA}}: This strategy trains neural networks, or parts of them, by means of adversarial learning in order to carry out the adaptation process between domains. In this category, \ac{GAN}~\cite{goodfellow2014generative} can be highlighted as a generative approach that is composed of a generator and a discriminator. The generator attempts to convert source images into target images, and the discriminator attempts to differentiate whether the image is part of the target or a fake. This tunes the weights of the network, signifying that the transformed source images cannot be differentiated from the target ones. Another relevant example is the Domain-Adversarial Neural Network~\cite{ganin2016domain}, a classification network based on the use of a \acf{GRL}. This network is divided into two parts that compete during training: one that learns to classify and the other that learns to distinguish the domain. \ac{GRL} is used to penalize the features that allow a distinction to be made between domains, such that only common features (called \textit{domain-invariant} features) are learned. This strategy was recently extended by Gallego \etal~\cite{gallego2020incremental} in order to incorporate a self-labeling incremental learning process that improves the results obtained.
\end{itemize}

Most of the \ac{DA} methods described above are focused principally on classification and are not directly applicable to binarization problems. In binarization, the method has to distinguish the relevant information from the rest of the content and has to return a response for each pixel of the image, rather than a single category.

\RevNote{A task related to binarization is that of image segmentation, which aims to recognize relevant elements from a given image by classifying them according to a set of categories. Some works have applied \ac{DA} to image segmentation. For example, Danbing \etal~\cite{ijcai2020-455} segments medical images using a combination of reconstruction and adversarial strategies based on the so-called CycleGAN~\cite{zhu2017unpaired}. Despite the good results reported, this solution was evaluated only in very similar domains (in fact, they belonged to the same dataset). Several approaches based on \ac{GAN} can also be found in the literature, such as the one proposed by Hoffman~\etal{}~\cite{pmlr-v80-hoffman18a}, which aims to obtain a common latent representation for the two domains involved, or the work by Yunsheng~\etal{}~\cite{li2019bidirectional}, whose proposal performs an image translation from source to target for then applying a segmentation image model. The \ac{GAN} and the segmentation model are trained through bidirectional learning, by connecting the outputs of each model to the input of the other. Also, Haq and Huang~\cite{haq2020adversarial} combined a \ac{GAN} model with an autoencoder to transform the target images so that a discriminator was not able to differentiate them, thus allowing both domains to be segmented with the same model. However, the goodness of these works was proved with domains in which the elements to be recognized keep their shape, thereby representing a very different situation to the binarization case.}

\RevNote{As regards the particularities of binarization, images only contain pixels categorized as background or ink; however, the ink class may represent different elements (such as text, musical notes, decorations, etc.) with different colors and shapes and with high-detail labeling. These characteristics depend on the nature of the documents, the engraving mechanisms or the different degradation levels associated with the course of time. These factors are hardly found in segmentation tasks, so they represent a significant challenge for methods such as GAN, in which it does not make sense to transform, for example, musical notes in handwritten text to binarize them, but it is rather necessary to look for other types of common features.}

Unlike existing approaches, we propose an unsupervised \ac{DA} method for document image binarization based on the state-of-the-art \ac{SAE} architecture and the \ac{GRL}. We extend the use of this layer to binarization tasks in order to process new domains without using labeled data. \RevNote{As introduced in~\secref{sec:introduction}, when the involved domains are alike, it could happen that using a model trained only with the source data performs better than with \ac{DA}. Given that the similarity between domains is, therefore, an important factor to be considered, we will propose a measure to evaluate the new target data before applying \ac{DA}.}

\subsection{\RevNote{Domain Similarity}}
\label{sec:sim}

\RevNote{Given its relevance in the present work, we now discuss the existing attempts to measure domain similarity. In general, DA techniques are based on the search for similarities or differences between domains. All the methods reviewed in the previous section include some form of comparison that seeks to eliminate the differences or bring together the similarities between the source and target domains. For example, the GRL~\cite{ganin2016domain} mentioned in the previous section is connected to a part of the network that performs the comparison of domains, which is used for learning the domain-invariant features.}

\RevNote{Many of these methods propose loss functions that calculate the similarity between domains \cite{sun2016deep, yan2017mind, bhushan2018deepjdot}, suggest strategies to obtain a common representation of the data in order to use the same classifier for both source and target domains \cite{ghifary2016deep, isola2017image}, or propose generative networks that analyze the similarities and differences of the domains in order to transform the images from source to target and vice versa~\cite{ijcai2020-455}. We can also find some proposals that include specific networks for calculating the similarity between domains, such as the Domain2Vec architecture proposed for classification tasks by Peng \etal~\cite{Peng2020Domain2Vec}, which makes use of a Siamese network to compare the domains, or the work by Osumi \etal~\cite{Osumi2019InstanceWeighting} which weights the contribution of the domain samples in the adaptation process by using their similarity. 
}

\RevNote{
In addition, it should be noted that in all these strategies, domain adaptation is always carried out, regardless of whether this process is adequate or not. However, if the domains are similar, there is also the possibility of using the model learned with the source domain to process the samples from the target domain. In this paper, we propose an external process that is carried out before the \ac{DA} task with the aim of determining whether the adaptation process is necessary. This initial check, in addition to achieving better results (as will be shown in the experimentation section), also means a general reduction in training time.}

\RevNote{
Our proposal to measure domain similarity (described in detail in \secref{sec:methodology:auto_da_bin}) is inspired by the \ac{IS} metric~\cite{Salimans2016IS} used to assess the quality of images created by \ac{GAN}. \ac{IS} uses a pre-trained network to classify the images generated and calculates a series of statistics from the probability distribution obtained, assuming that the images that represent objects in a clear and realistic way will activate certain classes with a higher probability, and that the images without objects or with unclear or unrealistic figures, will obtain a uniform distribution, in which none of the classes will stand out.
}

\RevNote{
Aligned to the idea of \ac{IS}, in this paper we propose a mechanism to determine the domain similarity by relying on the probability distribution provided by a neural network trained to binarize source images. Our hypothesis is that the activation obtained by this model for a new domain will allow us to compare its similarity with the domain used to train the network. That is, the network will provide similar responses for similar samples, and, therefore, when the domains are not alike, the distributions will not be alike either.
}

\RevNote{
It is also important to clarify that the proposed metric, rather than measuring the similarity between domains in a generic way, it compares the response of the network for a new domain with that given for the domain it was trained with. Hence, with all this, and supported by the idea of \ac{IS}, this metric is especially suitable for estimating whether the network will be successful with the new domain.
}

\section{Methodology}
\label{sec:methodology}

\subsection{Problem formulation}

Let \SourceDS{} be an annotated or \emph{source} dataset for document image binarization, composed of pairs in the form (\XSource{},~\YSource{}), where \XSource{}~$= [0, 255]^{h_{\text{s}}\times{}w_{\text{s}}\times{}c}$ is a document image of size $h_{\text{s}}\times{}w_{\text{s}}$~px and $c$ channels (for instance, $3$ in color and $1$ in grayscale), and \YSource{}~$= {\{0, 1\}}^{{h_{\text{s}}\times{}w_{\text{s}}}}$ is its corresponding pixelwise binary annotation.

Let \TargetDS{} be a non-annotated or \emph{target} dataset, which consists solely of a list of images \XTarget~$= {[0, 255]}^{h_{\text{t}}\times{}w_{\text{t}}\times{}c}$, with a size of $h_{\text{t}}\times{}w_{\text{t}}$ px and $c$ channels.

The task addressed in this paper is that of learning a model from \SourceDS{} and \TargetDS{}, with the aim of correctly binarizing images belonging to \TargetDS{}. Note that the overall problem is semi-supervised because it employs both labeled and unlabeled data; however, we refer to it as \emph{unsupervised DA} because we assume that there are no labeled data for the target set.

\subsection{\acl{SAE}}
\label{sec:methodology:sae}

The backbone of the method presented in this paper is an \ac{SAE}, which has been successfully used for document binarization in literature~\cite{CALVOZARAGOZA201937}. Given its importance in the present work, we introduce this architecture before integrating it into a \ac{DA} process.

An \ac{SAE} is a kind of neural network architecture that receives an image as input and provides another image with the same size as output but with values in the range of $[0,1]$. This network usually consists of two parts: an encoder, which processes the image through the use of consecutive pooling operators in order to extract meaningful features, and a decoder, which contains as many oversampling operators as pooling layers in the encoder with the objective of retrieving the original size of the image. 

However, the \ac{SAE} does not binarize directly, but obtains a probabilistic map on which the value of each pixel represents the probability of being foreground. Given an image \XSource{}, the \ac{SAE}, therefore, computes the map of probabilities \ProbSource{}, in order to subsequently apply a decision to each pixel, typically based on a probability threshold \ThresholdS{}.

Once the \ac{SAE} has been trained with a set of images, it can be used to process the remaining images from the same collection. In this case, since all the documents involved belong to the same domain, the performance is expected to be successful.

It should be kept in mind that the above approach splits the images into multiple patches of a fixed size, $h~\times~w$ px, during both training and inference. In the latter case, the binarized version of each patch is retrieved and then combined with the others in order to assemble the full binarized image.

\subsection{Domain adaptation for binarization}
\label{sec:methodology:dann}

One weak point of the \ac{SAE} is that of dealing with images belonging to different corpora with respect to that used in the training process. Since there is no labeled information for \TargetDS{}, the model can be trained only with~\SourceDS{}. Given the knowledge provided by \SourceDS{}, the \ac{SAE} model learns the features that allow any image~\XSource{} to be binarized. However, it will be able to successfully binarize images from \TargetDS{} if  \SourceDS{} and \TargetDS{} are similar domains, and the performance will be severely affected otherwise~\cite{CALVOZARAGOZA201937}.

In order to alleviate this problem, we propose an adversarial scheme based on the well-known \ac{GRL}~\cite{ganin2016domain}, which is a layer that penalizes those features that make it possible to distinguish the domains involved. This is useful as regards obtaining a domain-invariant representation that enables a successful binarization regardless of the domain.

\begin{figure}[!ht]
\centering
\includegraphics[width=1.0\textwidth]{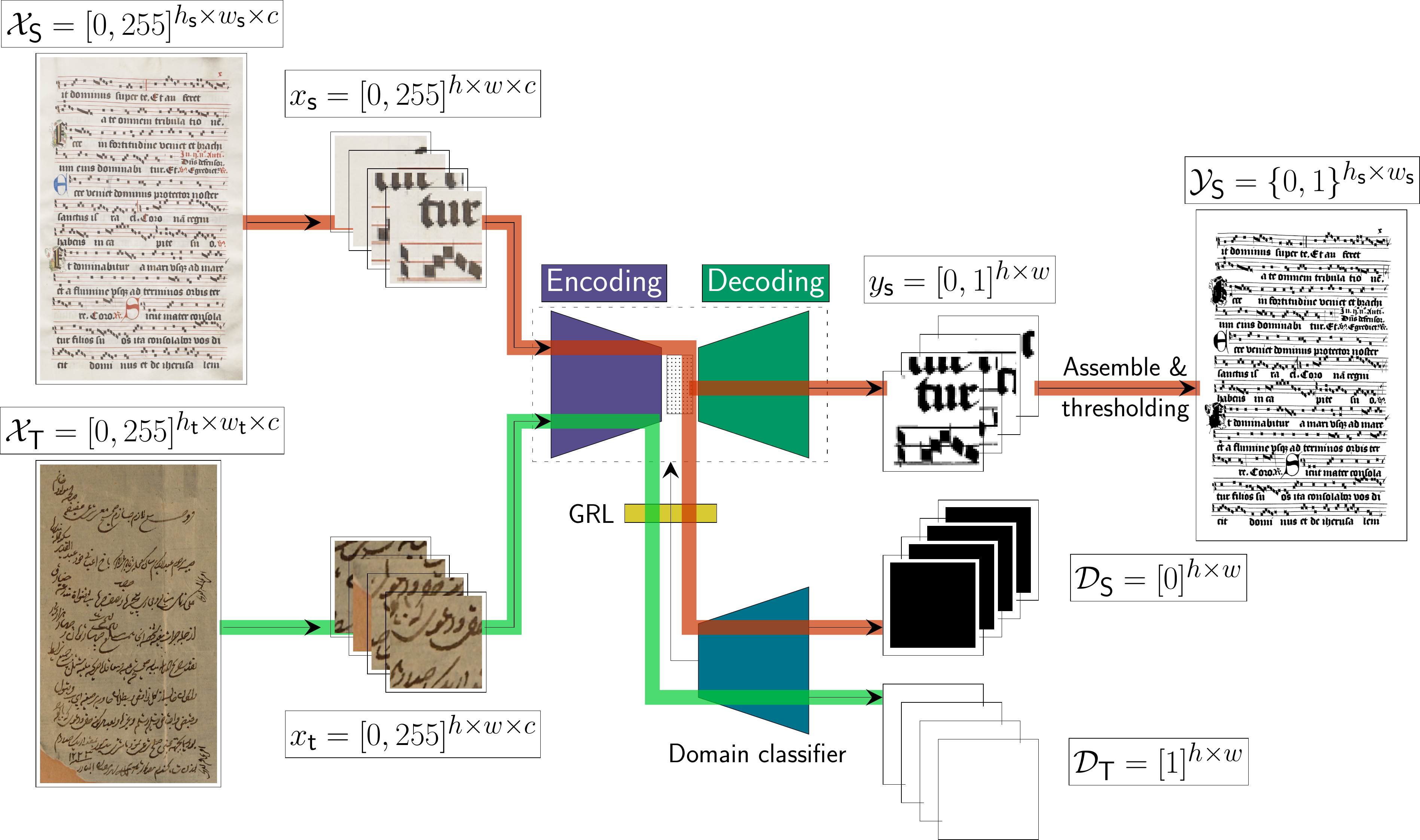}
\caption{Scheme of the neural domain adaptation approach proposed for document image binarization. The red line represents the path that the source images follow, while the green line represents the route of the target images. The classifier domain is trained in order to obtain images with all 0s or 1s, according to the origin of the image, i.e., from the source or target, respectively.}
\label{fig:scheme}
\end{figure}

Given \SourceDS{} and \TargetDS{}, defined previously, our approach must binarize pages from \TargetDS{} without any ground-truth from that domain. A graphic representation of the method that we propose is shown in~\figref{fig:scheme}. It consists of an \ac{SAE} architecture with which to binarize images, in which a \ac{GRL} connects one of its layers to a domain classifier that can differentiate the domain of the input image. \ac{GRL} assumes a hyper-parameter $\lambda$ in order to adjust the contribution of the domain classifier when training, which should be set empirically. 

With regard to the architecture of the domain classifier, it is important to stress that the different number of weights to be adjusted in both streams---binarization and domain classification---might yield an unbalanced structure during the training, thus causing both parts of the network to contribute in different scales to the tuning of the weights. In preliminary experiments, we observed that employing a simple architecture to discriminate the domain with a number of trainable parameters much lower than the binarization stream barely affects the overall training. This can be controlled by manually tweaking the parameter $\lambda$. However, a great difference between both parts of the network forces the tuning of this parameter to very specific values. We consequently decided to replicate part of the \ac{SAE} architecture for the domain classification in order to equal the number of weights in both branches.

\subsection{Adaptation applicability via domain similarity}
\label{sec:methodology:auto_da_bin}

When both the source and target domains are different, the supervised training from \SourceDS{} may not be sufficient to binarize images from \TargetDS{} reliably. In this situation, using the \ac{DA} process to adapt the learning of the target domain is a potential solution by which to binarize target images. However, the \ac{DA} process might not be appropriate when the \SourceDS{} and \TargetDS{} domains depict similar features. In this case, the GRL would be forced to forget useful features for binarization in both domains, and would pay attention to nuances in order to discriminate domains, which would be detrimental for the overall performance. It is for this reason that, in addition to adding \ac{DA} to the \ac{SAE} network, we consider it necessary to employ a strategy with which to determine when it is worth applying \ac{DA}. We, therefore, propose an \RevNote{unsupervised and} innovative mechanism that can be used to assess the similarity between \SourceDS{} and \TargetDS{}, in the context of binarization, in order to decide whether or not to eventually employ DA. 

\begin{figure*}[!ht]
\centering\includegraphics[width=1.\linewidth]{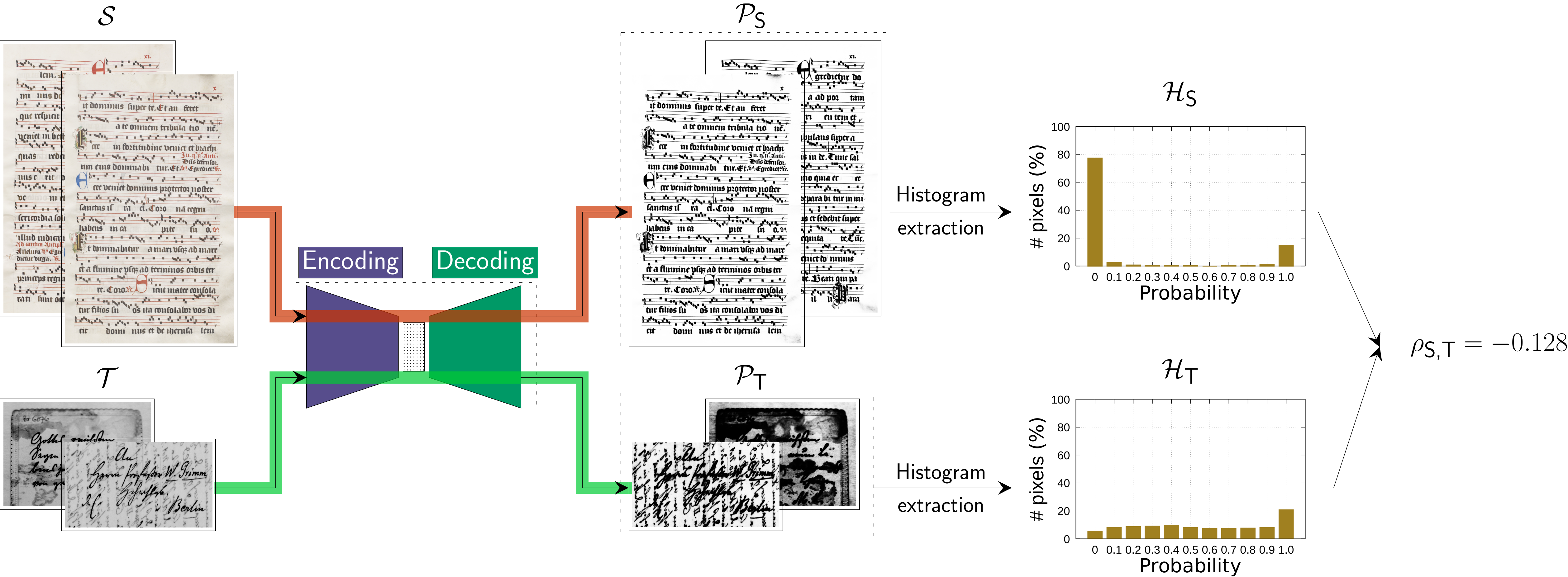}
\caption{Scheme of the domain similarity computation between two different corpora. The red path represents source processing, while the green path represents the route of the target images. The \ac{SAE} trained with \SourceDS{} predicts the binarization of both domains. The probabilistic map of each image is then used to obtain a global and normalized histogram of each domain in order to eventually calculate their correlation \RHOst{}. Note that the images are processed by means of patches, as explained in~\secref{sec:methodology:sae}.}
\label{fig:auto_scheme_example}
\end{figure*}

The scheme of our mechanism is shown in~\figref{fig:auto_scheme_example}. \RevNote{Since we are dealing with an unsupervised scenario, the target domain does not include ground-truth data to train the model or to evaluate the result. Instead, we will resort to the probability map obtained at the network's output, for which it is not necessary to use labels.} As explained previously, given any image from \SourceDS{}, the \ac{SAE} model \RevNote{(trained using only source images)} obtains a map \ProbSource{} showing the probability of each pixel being foreground. It is thus possible to process the images from \SourceDS{} and obtain their probability distribution \ProbSource{}. This model, when trained with \SourceDS{}, can similarly be used to compute the probability distribution \ProbTarget{} of the images from \TargetDS{}. 

Once the probability distributions of all the images from both domains (\ProbSource{} and \ProbTarget{}) have been attained, we obtain their normalized histograms (\HistSource{} and \HistTarget{}, respectively) by quantifying their probability figures with equitable intervals of \HistPrec{}. \RevNote{In other words, the SAE model provides the probability of each pixel to be ink or background, with values within the range [0, 1]. Our proposal maps this probability information onto a histogram for all the pages that belong to a domain. This histogram is a one-dimensional vector obtained by counting the number of pixels with the same probability values within a small range (precision or granularity \HistPrec{}$= 0.1$). Following this process, we generate two one-dimensional vectors, one for the source domain and one for the target domain, which represent the probability distribution histograms of these domains. Note that both are obtained with the \ac{SAE} trained with the labeled domain (source). Since the number of samples and their resolution can vary among different domains, these histograms are normalized according to the total number of pixels in each domain.}

\RevNote{We then compute the correlation~$\RHOst{}$ between these two normalized histograms. For this, we considered the Pearson's correlation coefficient~\cite{benesty2009pearson}. As we will show in \secref{sec:experiments:domain_similarity}, we experimented with other similar metrics and reached equivalent conclusions. We selected the Pearson's correlation coefficient because it is a linear correlation that can measure the similarity between two distributions symmetrically and with bounded values.} The following equation shows its mathematical definition

\begin{equation}
\rho{}_{S,T} = \dfrac{\text{cov}(\HistSource{}, \HistTarget{})}{\sigma{}_\text{S} \sigma{}_\text{T}},
\label{eq:correl}
\end{equation}

\noindent where $\text{cov}\left(\cdot\right)$ is the covariance function between two variables, while $\sigma{}_\text{S}$ and $\sigma{}_\text{T}$ represent the standard deviation of the histograms of \SourceDS{} and \TargetDS{}, respectively. Note that Eq.\ref{eq:correl} computes a similarity value between $-1$ and $1$, where $-1$ is employed for opposite distributions and $1$ for similar ones.

The coefficient $\RHOst{}$ provides a single value of source and target similarity. The decision to apply DA is made by using a threshold \RHOth{}, whose importance will be studied in greater detail in~\secref{sec:experiments}. This threshold divides the range of correlation values into two groups: when $\RHOst{}>\RHOth{}$, it is assumed that both domains are similar, and the \ac{SAE} model with no \ac{DA} is chosen; and when $\RHOst{}\leq{}\RHOth{}$, the domains involved can be considered as being different, and the adaptation strategy explained in Section \ref{sec:methodology:dann} is used. 

The entire process described in this section is detailed in~\algref{alg:pseudocode}, in which
$b$ and $e$ are the batch size and the number of epochs used to train the involved neural networks, respectively, and $h$ and $w$ are the height and width of the extracted patches of the documents, respectively; 
$\texttt{getHistogram}(\cdot)$ is a function that provides the histogram from a probabilistic map (\ProbSource{} or \ProbTarget{}) with a precision \HistPrec{}, 
and $\texttt{correlation}(\cdot)$ is another function with which to compute the linear correlation between two distributions using Pearson's coefficient.

\begin{algorithm}[!ht]
\DontPrintSemicolon
\KwIn{\SourceDS{} $\leftarrow$ $\{(\XSource{},~\YSource{})\}$ \newline \TargetDS{} $\leftarrow$ $\{(\XTarget{})\} \newline \lambda,~\RHOth{},~\HistPrec{},~h,~w,~e,~b~\leftarrow~\text{hyper-parameters}$}
\KwResult{$\mathcal{B}_{\text{T}}$~$\leftarrow$~Binarized images from \TargetDS{}.}
$\HistSource{}~\leftarrow~\varnothing$ \\
$\HistTarget{}~\leftarrow~\varnothing$ \\
$\ThresholdS{}~\leftarrow~\text{Fit SAE with}~\{\SourceDS{}, e, b, h, w\}$ \label{alg:trainSAE}\\
$\ProbTarget{} = \text{SAE prediction with}~\{(\XTarget{}, h, w)\}$\\
\ForEach{\XSource{}~\text{ in }~\SourceDS{}}{
$\ProbSource{} = \text{SAE prediction with}~\{(\XSource{}, h, w)\}$ \label{alg:predSAESource}\\
$\HistSource{}~\leftarrow~\HistSource{}~\cup~\texttt{getHistogram}(\ProbSource{}, \HistPrec{})$ \label{alg:histSource}\\
}
\ForEach{$\XTarget{}~\text{ in }~\TargetDS{}$}{
$\ProbTarget{} = \text{SAE prediction with}~\{(\XTarget{}, h, w)\}$ \label{alg:predSAETarget}\\
$\HistTarget{}~\leftarrow~\HistTarget{}~\cup~\texttt{getHistogram}(\ProbTarget{}, \HistPrec{})$ \label{alg:histTarget}\\
}
$\RHOst{} = \texttt{correlation}(\HistSource{}, \HistTarget{})$ \label{alg:correl}\\
\eIf{$\RHOst{}\leq{}\RHOth{}$}{
$\ThresholdS{}~\leftarrow~\text{Fit \BinDANN{} with}~\{\SourceDS{}, \TargetDS{}, e, b, h, w, \lambda\}$ \label{alg:trainDANN}\\
$\mathcal{B}_{\text{T}}~\leftarrow~\texttt{binarize with \BinDANN{}}(\TargetDS{}, h, w, \text{th}_\text{s})$ \label{alg:binDANN}
}{
$\mathcal{B}_{\text{T}}~\leftarrow~\texttt{binarize with SAE}(\TargetDS{}, h, w, \text{th}_\text{s})$ \label{alg:binSAE}
}

\Return $\mathcal{B}_{\text{T}}$
\caption{Adaptation applicability via domain similarity.}
\label{alg:pseudocode}
\end{algorithm}

The algorithm begins training the \ac{SAE} model with \SourceDS{} (\linealgref{alg:trainSAE}). After this training, it then stores a threshold \ThresholdS{}, with the best thresholding applied to the probabilistic map obtained with \SourceDS{}. This is computed in each epoch, validating the performance with different equidistant thresholds. That which optimizes the performance is then the threshold selected to be applied in binarization.

The next step is to obtain the probabilistic map \ProbSource{} of each image within \SourceDS{} by means of the \ac{SAE} model (\linealgref{alg:predSAESource}) and to compute an accumulative global histogram for that domain \HistSource{} (\linealgref{alg:histSource}). The same operation is performed for \TargetDS{}, obtaining the histogram \HistTarget{} from each \ProbTarget{} provided by the \ac{SAE} (lines~\ref{alg:predSAETarget},\ref{alg:histTarget}).

The similarity measure \RHOst{} is then computed between \HistSource{} and \HistTarget{} (\linealgref{alg:correl}) and compared with an input threshold \RHOth{}. If the correlation does not surpass the threshold, it is assumed that \SourceDS{} and \TargetDS{} are different domains, and it is, therefore, necessary to apply the \ac{DA} process. In this case, the domain adaptation model \RevNote{(represented as \BinDANN{})} is trained with \SourceDS{} and \TargetDS{}, and finally binarizes \TargetDS{} by using \ThresholdS{} (lines~\ref{alg:trainDANN},\ref{alg:binDANN}). In the opposite case, when \SourceDS{} and \TargetDS{} obtain high correlation, they are not considered as different domains, and \TargetDS{} is, therefore, binarized with the \ac{SAE} (\linealgref{alg:binSAE}).

Note that \SourceDS{} should be split into two partitions, for training and validation. The threshold \ThresholdS{} and the histogram \HistSource{} are adjusted with the validation partition.

\section{Experiments} 
\label{sec:experimental_setup}

In this section, we first describe the datasets and the metrics considered for evaluation purposes. We then detail the neural architecture used and the tuning of hyper-parameters, and finally, we present the results of the experiments and analyze them by comparing with those obtained with the state-of-the-art method. The experiments were implemented with the Keras v. 2.3.1 \citep{chollet2015keras} library and TensorFlow v. 1.14 as backend.

\subsection{Corpora}
\label{sec:corpora}

\begin{figure}[!ht]
\centering
\begin{subfigure}[b]{0.325\columnwidth}
\centering
\includegraphics[width=0.90\textwidth]{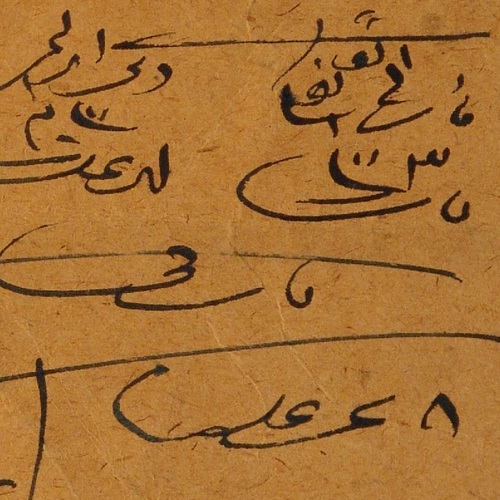}
\caption{\PhiD{}}
\end{subfigure}
\begin{subfigure}[b]{0.325\columnwidth}
\centering
\includegraphics[width=0.90\textwidth]{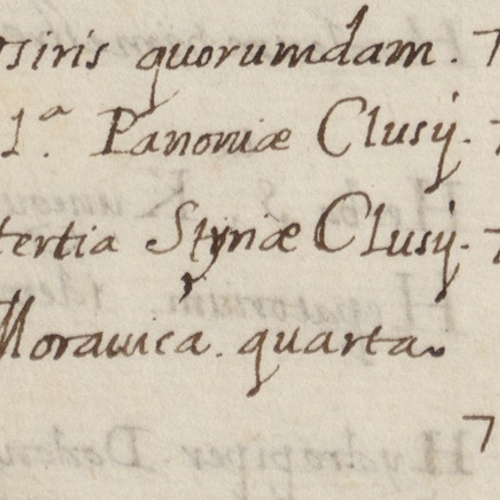}
\caption{\Dibco{}}
\end{subfigure}
\begin{subfigure}[b]{0.325\columnwidth}
\centering
\includegraphics[width=0.90\textwidth]{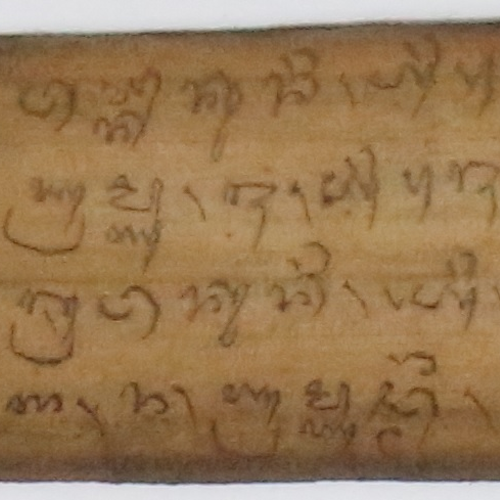}
\caption{\Palm{}}
\end{subfigure}

\begin{subfigure}[b]{0.325\columnwidth}
\centering
\includegraphics[width=0.90\textwidth]{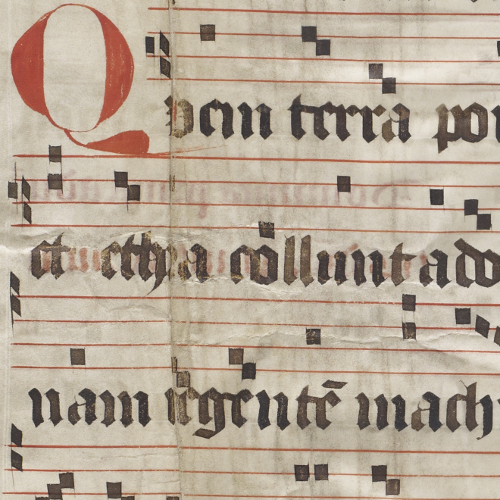}
\caption{\Salzinnes{}}
\end{subfigure}    
\begin{subfigure}[b]{0.325\columnwidth}
\centering
\includegraphics[width=0.90\textwidth]{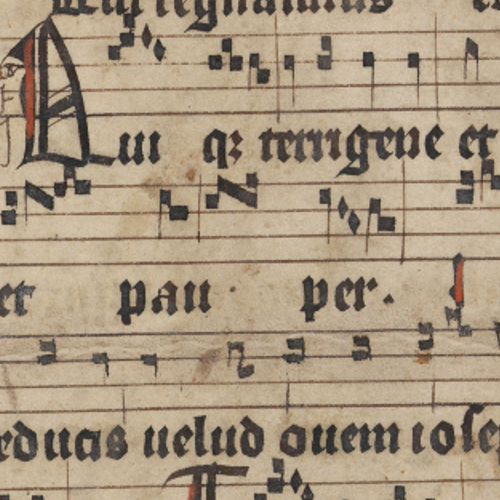}
\caption{\Einsiedeln{}}
\end{subfigure}

\caption{Some representative image regions from the corpora considered.}
\label{fig:examples_corpora}
\end{figure}

We evaluated our method by considering several datasets commonly used for document binarization (some examples can be seen in \figref{fig:examples_corpora}):
\begin{itemize}
\item \Dibco{}: the Document Image Binarization Contest has been held from 2009~\cite{gatos2009icdar}, and its datasets, containing different content, have been published each year. The experiments were carried out using the 2014 edition for the test set and the other editions until 2016 for the training set, as in \cite{CALVOZARAGOZA201937}. \RevNote{This corpus consists of 86 pages with an average size of $659\times{}1560$ px.}
\item \Einsiedeln{}: collection of 10 pages of mensural music documents, specifically those of Einsiedeln, Stiftsbibliothek, Codex 611(89)\footnote{\url{http://www.e-codices.unifr.ch/en/sbe/0611/}}. \RevNote{The images have an average size of $5550\times{}3650$ px.}
\item \Salzinnes{}: set of 10 pages of music score images of Salzinnes Antiphonal (CDM-Hsmu 2149.14)\footnote{\url{https://cantus.simssa.ca/manuscript/133/}} \RevNote{with $5100\times{}3200$ px., on average.}
\item \PhiD{}: dataset published for the Persian Heritage Image Binarization Competition (PHI) with a collection of Persian documents~\cite{ayatollahi2013persian}. \RevNote{It includes 15 pages with an average size of $1022\times{}1158$ px.}
\item \Palm{}: set of Balinese Palm Leaf manuscripts for the binarization competition organized at the 15th International Conference on Frontiers in Handwriting Recognition~\cite{burie2016icfhr2016}, which consists of 97 documents \RevNote{with a size of $492\times{}5116$ px., on average}, and whose ground truth was built by employing semi-automatic frameworks.
\end{itemize}

\subsection{Metrics}
\label{sec:metrics}
The binarization issue is a two-class problem in which pixels are classified into two possible classes: background and foreground. However, the amount of pixels belonging to each class is typically not balanced, thus promoting a possible bias towards the majority class (usually the background).

The metric of the binarization performance considered in this work was the {\it F-measure} (F$_{1}$). In a two-class classification task, this metric is defined as

\begin{equation}
\mbox{F}_{1} = \frac{2\cdot \mbox{TP} }{2 \cdot \mbox{TP} + \mbox{FP} + \mbox{FN}},
\label{eq:f1}
\end{equation}

\noindent where TP, FP, and FN represent {\it True Positives} or correctly classified elements, {\it False Positives} or type I errors, and {\it False Negatives} or type II errors, respectively. In the following, we consider the foreground as the \emph{positive} class.

\RevNote{To further analyze the performance, we also include results in terms of precision (\Precision{}) and recall (\Recall{}) metrics, defined mathematically as follows:}

\RevNote{
\begin{equation}
\mbox{P}= \frac{\mbox{TP}}{\mbox{TP} + \mbox{FP}}
\label{eq:precision}
\end{equation}
}

\RevNote{
\begin{equation}
\mbox{R}= \frac{\mbox{TP}}{\mbox{TP} + \mbox{FN}}
\label{eq:recall}
\end{equation}
}

\RevNote{Note that F$_{1}$ is the harmonic mean of these two metrics.}

\subsection{Impact of SAE and DA hyper-parameterization}
\label{sec:experiments}

Since an invariant-domain binarization method based on \ac{SAE} and \ac{GRL} is presented in this paper, it is necessary to study multiple parameters: those involved in the \ac{SAE} model and those associated with \ac{GRL}. In order to achieve a correct and robust configuration for our method, we first used three corpora---\Dibco{}, \Salzinnes{} and \PhiD{}---to subsequently assess its performance with the remaining collections. With regard to the \ac{SAE} proposed, in~\cite{CALVOZARAGOZA201937} it was demonstrated that, although the method attains high accuracy in on-domain scenarios, it is not robust in cross-domain ones. Since our proposal is based on that method, in this paper,  the experiments were carried out in order to compare them in terms of domain-adaptation ability. Note that the images are processed by the neural network in patches of $256 \times 256$ px, and that we considered 300 epochs for the training of all the models, keeping the best model according to a validation partition of the source dataset.

The \ac{SAE} considered in our experiments consists of an encoder with six convolution layers, with $64$ filters, kernels of $3\times3$, strides of $2\times2$, and \ac{ReLU} activations. The decoder performs the inverse operation with six transposed convolutional layers with the same configuration. After each \ac{ReLU} activation, a $0.2$ of dropout is performed for both the encoder and the decoder. We shall denote the set of convolution, \ac{ReLU} activation, and dropout layers as a \textit{block} of layers. After the last block of the decoder, we shall then apply a non-stride convolution layer with a sigmoid activation in order to obtain the probability of each pixel being foreground. Since residual connections from each encoding layer to its analogous decoding layer were a key factor in the original \ac{SAE} topology, our implementation also includes them. 

With regard to the integration with \ac{GRL}, the position of this layer within the \ac{SAE} architecture is relevant, since it affects only the previous layers to which it is connected. For instance, if \ac{GRL} is connected to the middle of the \ac{SAE} model, i.e., the so-called latent space of an auto-encoder, it will affect only the encoder section, and not that of the decoder (at least not directly). 

As mentioned in~\secref{sec:methodology:dann}, we propose an architecture in which the domain classifier and the binarization model contain a balanced number of trainable parameters.  We have, therefore, implemented the classifier by replicating part of the \ac{SAE} architecture, specifically with the same structure of the \ac{SAE} from the layer in which \ac{GRL} is connected. In preliminary experiments, we noticed that the model performed the binarization task better when the \ac{GRL} was connected before the last convolutional block of the decoder.  As regards the hyper-parameter $\lambda{}$ of GRL, we considered an incremental function that starts at $0.1$ and adds $0.01$ per epoch.

Since the model obtains a map of probabilities, thresholding is necessary in order to eventually determine which pixels belong to the foreground class and which belong to the background. However, it is supposed that only \SourceDS{} has available ground truth. For each epoch in training, we, therefore, calculate the best threshold, denoted as \ThresholdS{}, using a validation partition of \SourceDS{}. This threshold is then used to evaluate the images from \TargetDS{} in the experiments.

In the experiments, we first compare the baseline model, which we shall denote as SAE, with our \ac{GRL} approach, henceforth \BinDANN{}---, without, as yet, calculating the similarity domains. In these experiments, we considered all the possible combinations of pairs of datasets from the three selected. The pairs of datasets in the text were, for convenience, denoted as \SourceDS{}\StoT{}\TargetDS{} to refer to the combination of the labeled source dataset \SourceDS{} with the unlabeled target dataset \TargetDS{}.

The first set of results is reported in~\tabref{tab:preliminary}. This table shows that, despite the clear increase in accuracy achieved in several cases with respect to the state-of-the-art method, with an average performance of between $51.6\%$ and $66.9\%$ of F$_{1}$, not all the pairs of datasets considered in these preliminary experiments improve the baseline. This is particularly the case of the pairs \Dibco{}\StoT{}\Salzinnes{}, \PhiD{}\StoT{}\Salzinnes{} and \PhiD{}\StoT{}\Dibco{}, for which there is no such improvement. These results show that, for certain pairs of domains, it is not always better to perform the adaptation process, probably because of their similarity in the neural feature space. As we shall show later, thanks to the method proposed to measure this similarity, we shall improve the final result.

\begin{table}[!ht]
\caption{Preliminary experiments carried out for the three datasets selected \RevNote{in terms of precision, recall and F$_{1}$ (\%)}. The average row includes only the figures concerning the unsupervised scenarios. The \ac{SAE} \RevNote{columns represent} the state-of-the-art method, while the \BinDANN{} \RevNote{columns represent} the domain adaptation method without the comparison of histograms.}
\label{tab:preliminary}
\renewcommand{\arraystretch}{1.0}
\centering
\resizebox{1.0\textwidth}{!}{
\begin{tabular}{lll|ccccc:ccccc}
\hline
\multirow{2}{*}{\SourceDS{}}& \multirow{2}{*}{\TargetDS{}} & & \multicolumn{5}{c:}{SAE} & \multicolumn{5}{c}{\BinDANN{}} \\
    & & & & \Precision{} & \Recall{} & F$_{1}$ & & & \Precision{} & \Recall{} & F$_{1}$ & \\ \hline
\multirow{3}{*}{\Salzinnes{}} & \Salzinnes{} & & & 96.8 & 94.9 & 95.9 & & & - & - & - & \\ \cdashline{2-13} 
    & \Dibco{} & & & 60.1 & 96.2 & 74.0 & & & 74.8 & 93.3 & \textbf{83.0} & \\
    & \PhiD{} & & & 10.8 & 99.9 & 19.5 & & & 53.6 & 99.7 & \textbf{69.7} & \\ \hline
\multirow{3}{*}{\Dibco{}} & \Dibco{} & & & 94.7 & 83.4 & 88.7 & & & - & - & - & \\ \cdashline{2-13}
    & \Salzinnes{} & & & 99.8 & 78.6 & \textbf{88.0} & & & 80.2  & 69.1 & 74.2 & \\
    & \PhiD{} & & & 13.3 & 99.8 & 23.5 & & & 64.0 & 98.4 & \textbf{77.6} & \\ \hline
\multirow{3}{*}{\PhiD{}} & \PhiD{} & & & 85.5 & 83.5 & 84.5 & & & - & - & - & \\ \cdashline{2-13}
    & \Salzinnes{} & & & 99.9 & 40.4 & \textbf{57.5} & & & 82.8 & 37.3 & 51.4 & \\
    & \Dibco{} & & & 99.2 & 30.8 & \textbf{47.0} & & & 88.7 & 30.4 & 45.3 & \\ \hline
\multicolumn{3}{l|}{Average} & & 63.9 & 74.3 & 51.6 & & & 74.0 & 71.3 & \textbf{66.9} & \\ \hline
\end{tabular}
}
\end{table}

\RevNote{
Besides, we observe some differences between the \Precision{} and \Recall{} figures. The reason behind this is that the binarization method is based on a threshold set to the probability map provided by \ac{SAE}, which is calculated using only the images from \SourceDS{}. For example, for \Salzinnes{}\StoT{}\PhiD{} and \Dibco{}\StoT{}\PhiD{}, \ac{SAE} presents a severe reduction in the \Precision{} obtained. However, the adaptation process carried out by \BinDANN{} manages to improve this result. In other cases, we observe a high \Precision{} but a low \Recall{}. This is also associated with the threshold set in an unsupervised manner. In these cases, binarization provides few false positives, but some false negatives, mainly because the method classifies as ink only when the probability of being ink is very high. These results are also a symptom that the SAE is not working well with the target images, which, as we have indicated, is solved after the adaptation process.
}

\RevNote{
After analyzing these preliminary results, we observe that F$_{1}$, the harmonic mean of \Precision{} and \Recall{}, is enough for the evaluation. Therefore, from here on we will report the results considering only this metric. 
}

\subsection{Impact of domain similarity hyper-parameterization}
\label{sec:experiments:domain_similarity}

We shall now evaluate the proposed method with which to calculate the similarity between domains. As mentioned previously, it is necessary to establish a threshold \RHOth{} for the correlation value \RHOst{} in order to decide whether the model should be binarized with the supervised \ac{SAE} or with the model based on \ac{DA}. \RevNote{For the following experiments, we are going to consider the quotient $\nicefrac{\text{F}_1^{\BinDANN{}}}{\text{F}_1^{\text{SAE}}}$ as a measure of the relative improvement obtained by the DA process with respect to the conventional SAE.}

\RevNote{As mentioned in \secref{sec:methodology:auto_da_bin}, in addition to the Pearson's correlation, we considered a set of known metrics to compute the similarity or divergence between the two histograms \HistSource{} and \HistTarget{}. Specifically, we compared the Pearson's correlation~\cite{benesty2009pearson}, the Kullback–Leibler (KL) divergence~\cite{shlens2014notes}, the Jensen-Shannon (JS) divergence~\cite{briet2009properties} and the histogram intersection~\cite{lee2005evaluation}.}

\begin{figure}[!ht]
\centering\includegraphics[width=0.58\linewidth]{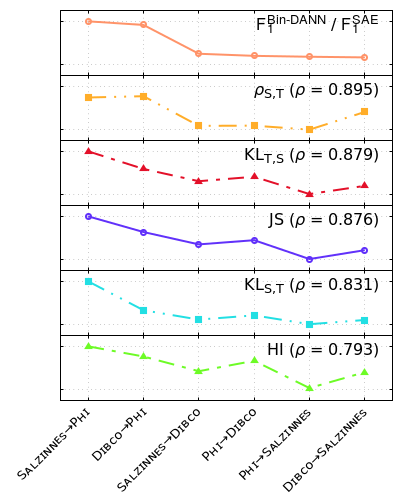}
\caption{\RevNote{Comparison among different similarity/divergence metrics between \HistSource{} and \HistTarget{}. The horizontal axis indicates the different pairs of domains \SourceDS{}\StoT{}\TargetDS{} studied. The first curve stands for the quotient $\nicefrac{\text{F}_1^{\BinDANN{}}}{\text{F}_1^{\text{SAE}}}$. The rest of graphs represent the different correlation/divergence metrics normalized between 0 and 1 for easy comparison. They are ordered according to the correlation ($\rho$) of each graph compared with the first curve, and whose values are included between brackets.}}
\label{fig:correl_pre}
\end{figure}

\RevNote{\figref{fig:correl_pre} shows a comparison among all these metrics, where it is observed that all have a similar trend. We also compared this trend with the curve obtained by the relative improvement $\nicefrac{\text{F}_1^{\BinDANN{}}}{\text{F}_1^{\text{SAE}}}$ for each pair of domains \SourceDS{}\StoT{}\TargetDS{}, ordered by this quotient. For further analysis, the figure includes the Pearson's correlation between each curve compared with the degree of improvement of \BinDANN{} with respect to SAE (i.e., the first curve). 
}

\RevNote{We observe that Pearson has the highest correlation, with a figure of $0.895$. The JS and KL divergences also obtain competitive figures between $0.879$ and $0.831$, whereas HI reports the worst correlation with $0.793$. It should be noted that KL divergence is not symmetric, so it changes according to the order in which the operations are computed. Particularly, we observe a correlation of $0.879$ for KL$_{\text{T,S}}$, and a value of $0.831$ for the opposite KL$_{\text{S,T}}$. For all this, we finally selected the Pearson's correlation for the rest of experiments, since in addition to reporting the best results, it also returns bounded values between -1 and 1, simplifying the search for the optimal threshold \RHOst{}. Anyway, since similar behaviors are observed for all cases, we can assume that our method is robust regardless of the particular metric used to compare the probability distributions.}

\begin{figure}[!ht]
\centering\includegraphics[width=0.75\linewidth]{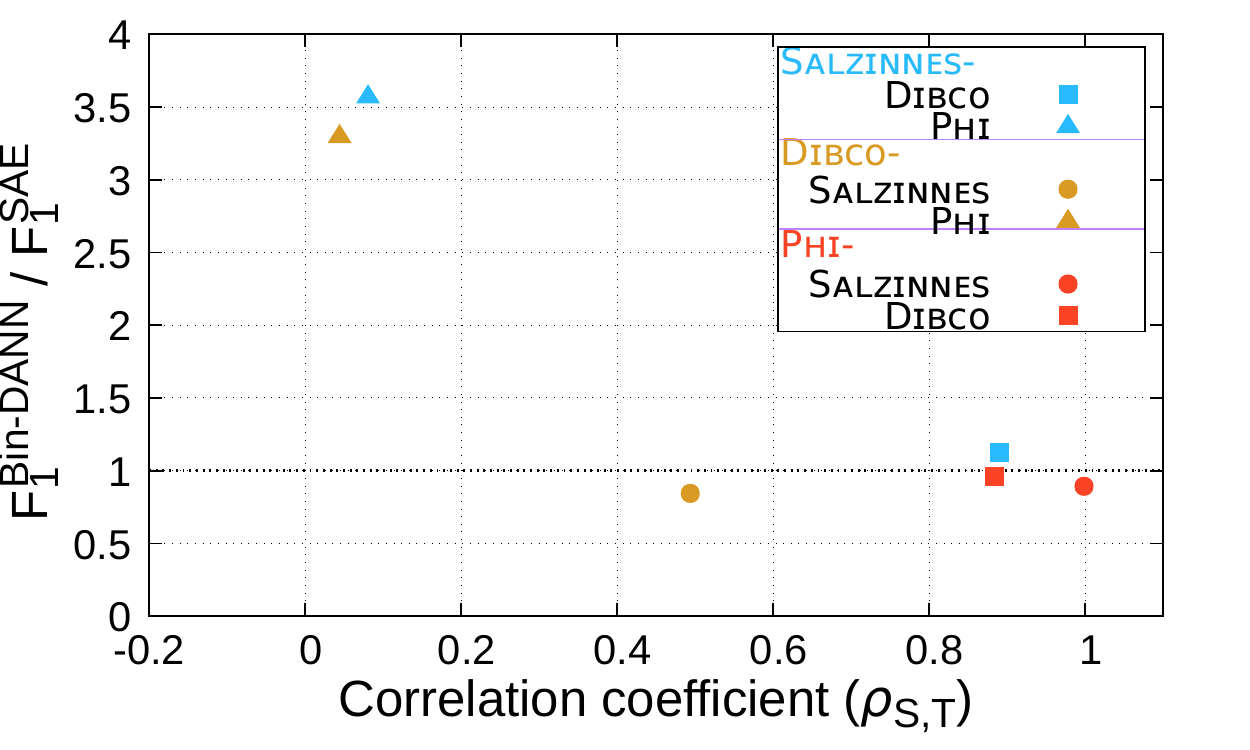}
\caption{Relationship between the correlation coefficient and the improvement obtained when applying domain adaptation. The horizontal dotted line marks the boundary of no improvement with respect to the \ac{SAE} method. Higher figures are those cases in which \BinDANN{} improves the results and the lower ones represent the cases in which it does not.}
\label{fig:correl_bindann_sae}
\end{figure}

In~\figref{fig:correl_bindann_sae}, we plot \RevNote{the quotient $\nicefrac{\text{F}_1^{\BinDANN{}}}{\text{F}_1^{\text{SAE}}}$} with respect to \RHOst{} in order to assess their relationship. This figure shows that when \RHOst{} is near to the maximum value, the improvement ratio obtained with \BinDANN{} is virtually non-existent, or even detrimental (below $1$). This phenomenon would appear to be associated with those cases in which the \ac{SAE} model provides similar histogram distributions for both the source and target domains. In these cases, and following our premise, the binarization can be performed by the \ac{SAE} trained with \SourceDS{}.

However, when the histogram distributions between \SourceDS{} and \TargetDS{} barely match (left-hand side of the horizontal axis), the improvement rate rises drastically, increasing to a maximum factor of over $3$ times with respect to the \ac{SAE} model. For example, in the case \Salzinnes{}\StoT{}\PhiD{}, we obtain an \ac{SAE} binarization performance of $19.5\%$, which is considerably improved by \BinDANN{} to $69.7\%$, or in the case \Dibco{}\StoT{}\PhiD{}, when the performance is increased from $23.5\%$ to $77.6\%$.

\begin{figure}[!ht]
\centering\includegraphics[width=0.75\linewidth]{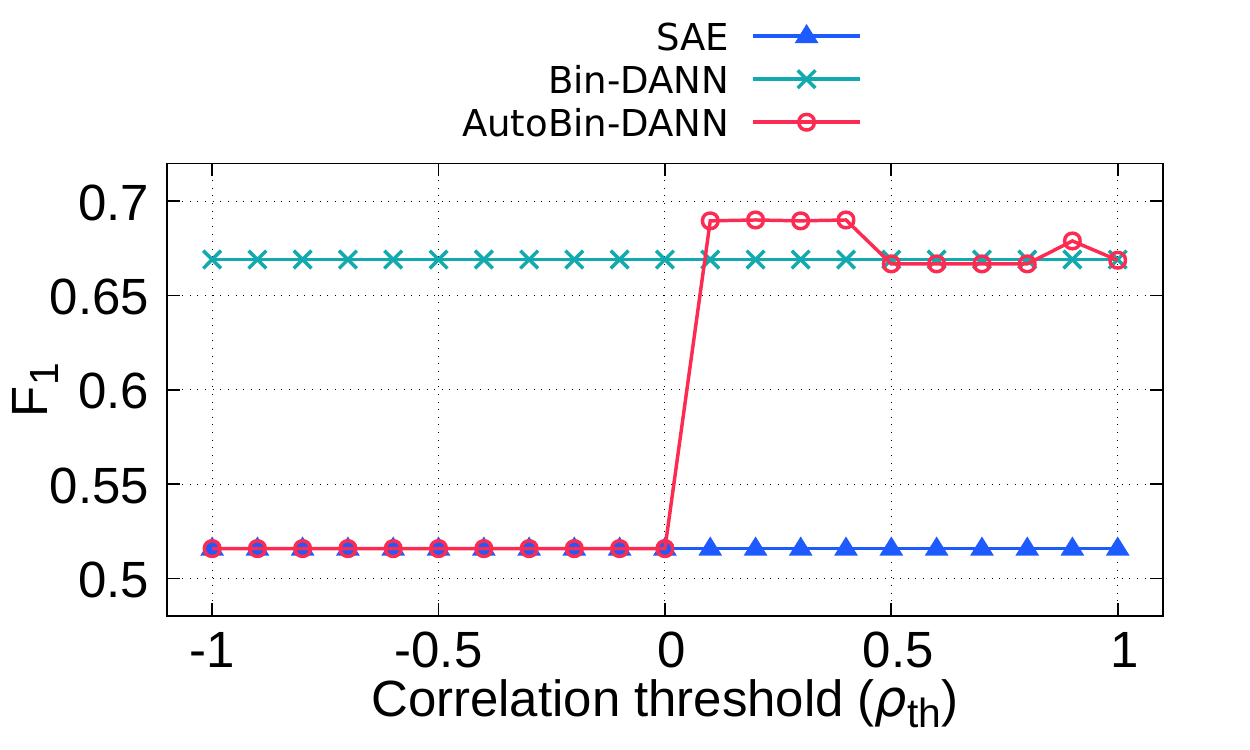}
\caption{Comparison between the averurlage performance obtained by \ac{SAE}, \BinDANN{} and \AutoBinDANN{}, depending on the threshold selected \RHOth{}. When the correlation \RHOst{} is less than or equal to \RHOth{}, the method selects \BinDANN{}; otherwise, it selects \ac{SAE}.}
\label{fig:correl_threshold}
\end{figure}

We shall now evaluate the method that automatically selects the \ac{SAE} or the \BinDANN{},  henceforth denoted as \AutoBinDANN{} (Algorithm \ref{alg:pseudocode}). We assess its performance with respect to \RHOth{} in~\figref{fig:correl_threshold}. The curves reveal a clear increase in performance of \AutoBinDANN{} when $\RHOth{}\geq{}0.1$. 
The best results are obtained in the range between $0.1$ and $0.4$, obtaining an F$_{1}$ value of $69\%$ for \AutoBinDANN{} when compared with the \ac{SAE} and the \BinDANN{}, with only $51.6\%$ or $66.9\%$, respectively.

The preliminary experiments, therefore, show that the best results are attained when $\RHOst{}\in[0.1, 0.4]$. For the final experiments, which are described in the next section, we consider the middle value within that range, i.e., $\RHOth{} = 0.25$.

\begin{table}[!ht]
\caption{\RevNote{Analysis of the similarity between images of the same domain \RHOss{}. We report the values in terms of Pearson's correlation (average $\pm$ std. deviation).}}
\label{tab:on-domain-correlation}
\renewcommand{\arraystretch}{0.98}
\centering
\resizebox{0.48\textwidth}{!}{
\RevNote{
\begin{tabular}{lc}
\hline
Domain & \begin{tabular}[c]{@{}c@{}}Correlation \RHOss{}\end{tabular} \\ \hline
\Salzinnes{}  & $0.99\pm 6 \times {\rm 10}^{-8}$ \\
\Dibco{}      & $0.95\pm 7 \times {\rm 10}^{-2}$ \\
\PhiD{}       & $0.99\pm 3 \times {\rm 10}^{-4}$ \\ \hline
\end{tabular}
}
}
\end{table}

\RevNote{To complement these experiments, \tabref{tab:on-domain-correlation} analyzes the values of the Pearson's correlation coefficient (\RHOss{}) when comparing samples within the same domain. We observe that, in all cases, the correlation is higher than 0.9 with very low variance. It must be kept in mind that, in these datasets, there are samples in which the color distribution is maintained but the ink density varies, as well as samples of the opposite. However, the values from \tabref{tab:on-domain-correlation} indicate that the intra-domain correlation is consistent and, therefore, the proposed metric is suitable for the task at issue.
}

\subsection{Final evaluation}
\label{sec:final_evaluation}

We shall now employ the best configuration determined previously to study the results obtained for all the combinations of the datasets considered. The first columns of \tabref{tab:full_results} show the F$_{1}$ obtained for the \ac{SAE}, \BinDANN{}, and \AutoBinDANN{} methods. We then include the correlation calculated for each combination of datasets, and finally, the improvement achieved with respect to the performance of \ac{SAE} when applying \BinDANN{} and \AutoBinDANN{}. 

As expected, the domain adaptation method (\BinDANN{}) is not always the best option, since there are several combinations of \SourceDS{} and \TargetDS{} that do not improve the performance of the \ac{SAE}. If we focus on the correlation column \RHOst{}, we can observe that higher figures of this metric, i.e., those near to $1$, are associated with both high binarization performance with \ac{SAE}, a bare improvement or a decrease in the quality of the binarization in the adaptation model.

\begin{table*}[!ht]
\caption{Final results for all corpora. The best results between the F$_{1}^\text{SAE}$, F$_{1}^\text{\BinDANN{}}$ and F$_{1}^\text{\AutoBinDANN{}}$ columns are highlighted in bold type. The \RHOst{} column is the correlation coefficient between the histograms obtained from the probabilistic map of \SourceDS{} and \TargetDS{} by means of the \ac{SAE} model. Finally, the two last columns show the difference between our proposals (\BinDANN{} and \AutoBinDANN{}) and the state-of-the-art model (\ac{SAE}).}
\label{tab:full_results}
\renewcommand{\arraystretch}{0.98}
\centering
\resizebox{0.95\textwidth}{!}{
\begin{tabular}{ll:ccc:c:cc}
\hline
\SourceDS{}                      & \TargetDS{}     & \begin{tabular}[c]{@{}c@{}}F$_{1}^\text{SAE}$\\ (\%)\end{tabular} & {\begin{tabular}[c]{@{}c@{}}F$_{1}^\text{\BinDANN{}}$\\ (\%)\end{tabular}} & {\begin{tabular}[c]{@{}c@{}}F$_{1}^\text{\AutoBinDANN{}}$\\ (\%)\end{tabular}} & \RHOst{} & {\begin{tabular}[c]{@{}c@{}}F$_{1}^\text{\BinDANN{}}$ \\- F$_{1}^\text{SAE}$ \end{tabular}} & {\begin{tabular}[c]{@{}c@{}}F$_{1}^\text{\AutoBinDANN{}}$ \\- F$_{1}^\text{SAE}$ \end{tabular}}\\ \hline
\multirow{5}{*}{\Salzinnes{}}  
& \Salzinnes{}  & 95.9 & - & - & \phantom{$-$}$1.00$ & \phantom{00}- & \phantom{00}- \\
& \Einsiedeln{} & 67.3 & \textbf{92.1} & 67.3 & \phantom{$-$}$0.60$ & $+24.8$ & \phantom{00}$0.0$ \\
& \Dibco{} & 74.0 & \textbf{83.0} & 74.0 & \phantom{$-$}$0.89$ & \phantom{0}$+9.1$ & \phantom{00}$0.0$ \\
& \PhiD{} & 19.5 & \textbf{69.7} & \textbf{69.7} & \phantom{$-$}$0.08$ & $+50.2$ & $+50.2$ \\
& \Palm{} & 12.1 & \textbf{32.7} & \textbf{32.7} & \phantom{$-$}$0.07$ & $+20.6$ & $+20.6$ \\ 
\hline
\multirow{5}{*}{\Einsiedeln{}}
& \Salzinnes{} & \textbf{90.4} & 88.2 & \textbf{90.4} & \phantom{$-$}$0.54$ & \phantom{0}$-2.2$ & \phantom{00}$0.0$ \\
& \Einsiedeln{} & 92.5 & - & - & \phantom{$-$}$1.00$ & \phantom{00}- & \phantom{00}- \\
& \Dibco{} & \textbf{86.1} & 84.2 & \textbf{86.1} & \phantom{$-$}$0.99$ & \phantom{0}$-1.9$ & \phantom{00}$0.0$ \\
& \PhiD{} & 20.8 & \textbf{63.0} & \textbf{63.0} & $-0.24$ & $+42.3$ & $+42.3$ \\
& \Palm{} & 16.0 & \textbf{30.6} & \textbf{30.6} & \phantom{$-$}$0.25$ & $+14.6$ & $+14.6$ \\ 
\hline
\multirow{5}{*}{\Dibco{}}  
& \Salzinnes{}  & \textbf{88.0} & 74.2 & \textbf{88.0} & \phantom{$-$}$0.49$ & $-13.7$ & \phantom{00}$0.0$ \\
& \Einsiedeln{} & \textbf{89.6} & 83.2 & \textbf{89.6} & \phantom{$-$}$0.99$ & \phantom{0}$-6.5$ & \phantom{00}$0.0$ \\
& \Dibco{}  & 88.7 & - & - & \phantom{$-$}$1.00$ & \phantom{00}- & \phantom{00}- \\
& \PhiD{} & 23.5 & \textbf{77.6} & \textbf{77.6} & \phantom{$-$}$0.04$ & $+54.1$ & $+54.1$ \\
& \Palm{} & 15.8 & \textbf{30.4} & \textbf{30.4} & \phantom{$-$}$0.13$ & $+14.6$ & $+14.6$ \\ 
\hline
\multirow{5}{*}{\PhiD{}}        
& \Salzinnes{} & \textbf{57.5} & 51.4 & \textbf{57.5} & \phantom{$-$}$0.99$ & \phantom{0}$-6.2$ & \phantom{00}$0.0$ \\
& \Einsiedeln{} & \textbf{62.2} & 60.2 & \textbf{62.2} & \phantom{$-$}$0.77$ & \phantom{0}$-2.0$ & \phantom{00}$0.0$ \\
& \Dibco{}  & \textbf{47.0} & 45.3 & \textbf{47.0} & \phantom{$-$}$0.88$ & \phantom{0}$-1.7$ & \phantom{00}$0.0$ \\
& \PhiD{} & 84.5 & - & - & \phantom{$-$}$1.00$ & \phantom{00}- & \phantom{00}- \\
& \Palm{} & \textbf{21.8} & 20.1 & \textbf{21.8} & \phantom{$-$}$0.92$ & \phantom{0}$-1.7$ & \phantom{00}$0.0$ \\ 
\hline
\multirow{5}{*}{\Palm{}}       
& \Salzinnes{} & 70.8 & \textbf{76.1} & 70.8 & \phantom{$-$}$0.32$ & \phantom{0}$+5.2$ & \phantom{00}$0.0$ \\
& \Einsiedeln{} & 73.8 & \textbf{80.7} & 73.8 & \phantom{$-$}$0.43$ & \phantom{0}$+6.9$ & \phantom{00}$0.0$ \\
& \Dibco{}  & 58.2 & \textbf{68.7} & 58.2 & \phantom{$-$}$0.66$ & $+10.5$ & \phantom{00}$0.0$ \\
& \PhiD{} & \textbf{66.6} & 19.5 & \textbf{66.6} & \phantom{$-$}$0.96$ & $-47.0$ & \phantom{00}$0.0$ \\
& \Palm{} & 39.4 & - & - & \phantom{$-$}$1.00$ & \phantom{00}- & \phantom{00}- \\ 
\hline
\end{tabular}
}
\end{table*}

In general, the performance of the \ac{SAE} is severely reduced when $\SourceDS{}\neq\TargetDS{}$. For example, if we focus on \Salzinnes{} as a source, we observe a great difference in performance as regards the on-domain experiment (\Salzinnes{} \StoT{}\Salzinnes{})---with a result of $95.9\%$---and the cross-domain ones, with figures of between $12.1\%$ and $74\%$. The same applies when \PhiD{} is used as a source: the on-domain experiment attains $84.5\%$, whereas the performance in the cross-domain experiments ranges from $21.8\%$ to $62.2\%$. 

When $\SourceDS{} \equiv \Dibco{}$, the results of the music datasets (\Salzinnes{} and \Einsiedeln{}) are also comparable with the on-domain case, with values near to $89\%$. This may be because \Salzinnes{} and \Einsiedeln{}, in addition to other elements, also contain text. Indeed, in the opposite situation, i.e., when \Salzinnes{}\StoT{}\Dibco{} and \Einsiedeln{}\StoT{}\Dibco{}, the performance decreases slightly. We attribute this phenomenon to the wide variety of elements in music manuscripts, such as staff lines, ornaments and music notes, which do not appear in \Dibco{}. 

With regard to the experiments with $\SourceDS{} \equiv \Palm{}$, the on-domain results are an exception. It is worth noting that this dataset contains poor-quality images, and this dataset does not, therefore, achieves good results, regardless of the \SourceDS{} considered, with the best case being the on-domain situation with $39.4\%$, when compared to the cross-domain experiments with values of only $21.8\%$ (\PhiD{}\StoT{}\Palm{}).

It should also be noted that \BinDANN{} can greatly increase the performance of binarization in many cases, while the correlation-based selection allows the baseline performance to be maintained when the situation is not suitable for adaptation. Furthermore, there are cases in which this decision is not the best. For example, in the case of \Salzinnes{}\StoT{}\Einsiedeln{}, the performance of \BinDANN{} improves from $67.3\%$ to $92.1\%$, but the correlation coefficient reaches a high value of $0.60$. It is precisely for this reason that the substantial improvement made by the domain adaptation method is not selected when the correlation threshold is set to $0.25$. This is not, of course, detrimental when compared to the baseline case, signifying that the binarization can be carried out with fair results. A similar situation can be found in the case of \Salzinnes{}\StoT{}\Dibco{}, with an improvement from $74.0\%$ to $83.0\%$, and \Palm{}\StoT{}\Dibco{}, from $58.2\%$ to $68.7\%$. 

In spite of the above, the \AutoBinDANN{} is beneficial in a number of cases. For example, when \Einsiedeln{}\StoT{}\PhiD{} or \Einsiedeln{}\StoT{}\Palm{}, the binarization results of \BinDANN{} are improved from $20.8\%$ to $63.0\%$ and from $12.1\%$ to $32.7\%$, respectively, and the correlation allows the eventual selection of the adapted method. There are similar cases for almost all the source scenarios, with the exception of \PhiD{}, since when it is used as a source, the adaptation is detrimental in all the target cases. However, the \AutoBinDANN{} manages to correct this issue by selecting the state-of-the-art method, and does not, therefore, compromise the binarization performance. 

After the correlation thresholding, none of the $20$ unsupervised experiments (without including those when the same domain is considered as source and target) obtained a loss of performance with respect to the \ac{SAE} model. The low threshold value obtained in the preliminary experiments makes it possible to increasing the robustness of \AutoBinDANN{}, but, as mentioned previously, there are cases in which the \ac{DA} approach leads to improvement, but which are not exploited owing to the correlation bias. 

Moreover, note that the adaptation process is not suitable when the performance of the baseline is high. For example, in the case of \Einsiedeln{}\StoT{}\Dibco{}, \ac{SAE} attains $86.1\%$, while \BinDANN{} obtains $84.2\%$, \Dibco{}\StoT{}\Salzinnes{} decreases from $88.0\%$ for \ac{SAE} to $74.2\%$ with \ac{DA} and \Dibco{}\StoT{}\Einsiedeln{} worsens the binarization from $89.6\%$ to $83.2\%$. We attribute this loss to the fact that the feature representation learned by \ac{SAE} works properly with \TargetDS{}, but \ac{DA} modifies and blurs it in the learning process. Furthermore, when \SourceDS{} is \Palm{}, the domain adaptation is slightly better for each target except for the combination \Palm{}\StoT{}\PhiD{}, in which it falls from $66.5\%$ to $19.5\%$. However, the correlation bias makes it possible to avoid this drawback by selecting the \ac{SAE} method. In addition, when \PhiD{} is used as a source, the domain adaptation is slightly detrimental for all the targets considered. \RHOth{}, therefore, plays an important role as regards avoiding these situations in order to maximize the robustness of the combined approach.

\begin{figure}[!ht]
\centering\includegraphics[width=0.75\linewidth]{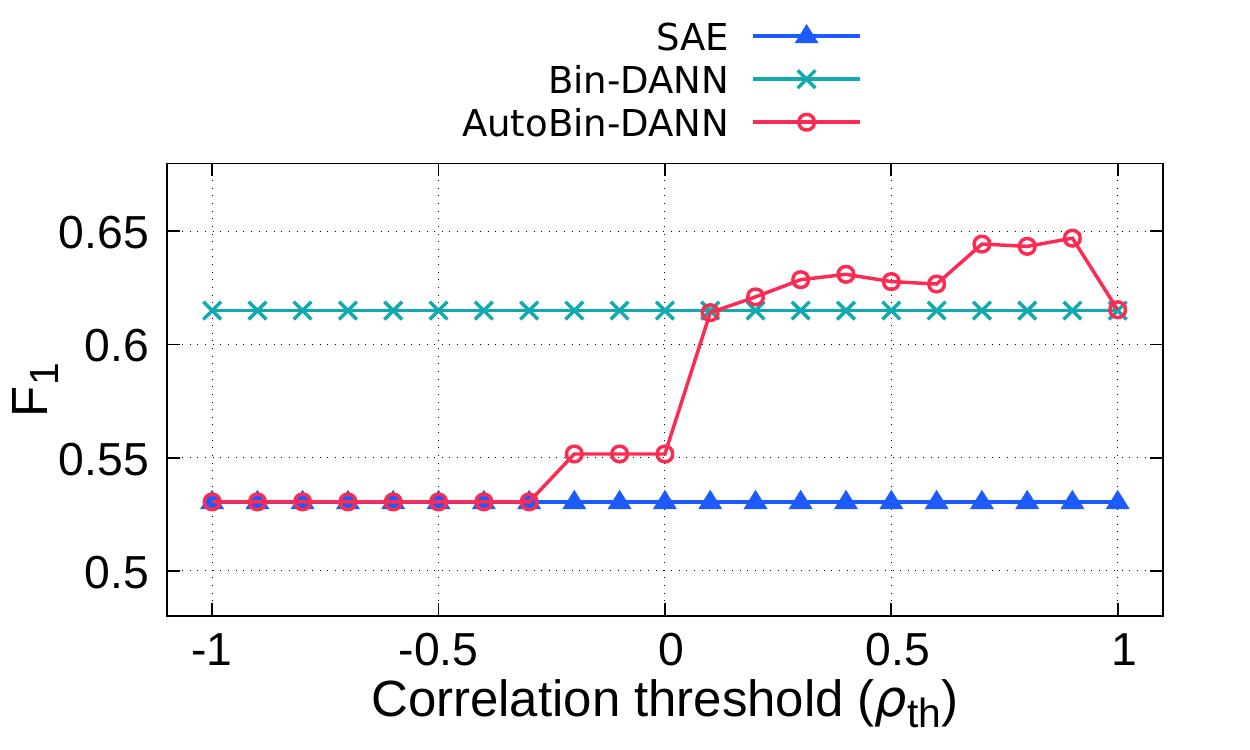}
\caption{Comparison between \ac{SAE}, \BinDANN{} and \AutoBinDANN{} depending on the threshold \RHOth{} selected.}
\label{fig:correl_bindann_sae_full}
\end{figure}

All  of the above will now be employed to analyze the importance of the correlation threshold in~\figref{fig:correl_bindann_sae_full}. Similar to what occurred in the preliminary analysis, \AutoBinDANN{} improves the \ac{SAE} model with $\RHOth{} \in [-0.2, 1.0]$, with this improvement being more relevant within the range $[0.1, 1.0]$. In these results,  the same trend is obtained with the datasets used for the adjustment of the hyper-parameters, in which, for thresholds from $0.1$, the model provided better results when applying the proposed \AutoBinDANN{}. 

\begin{figure}[!ht]
\centering\includegraphics[width=0.75\linewidth]{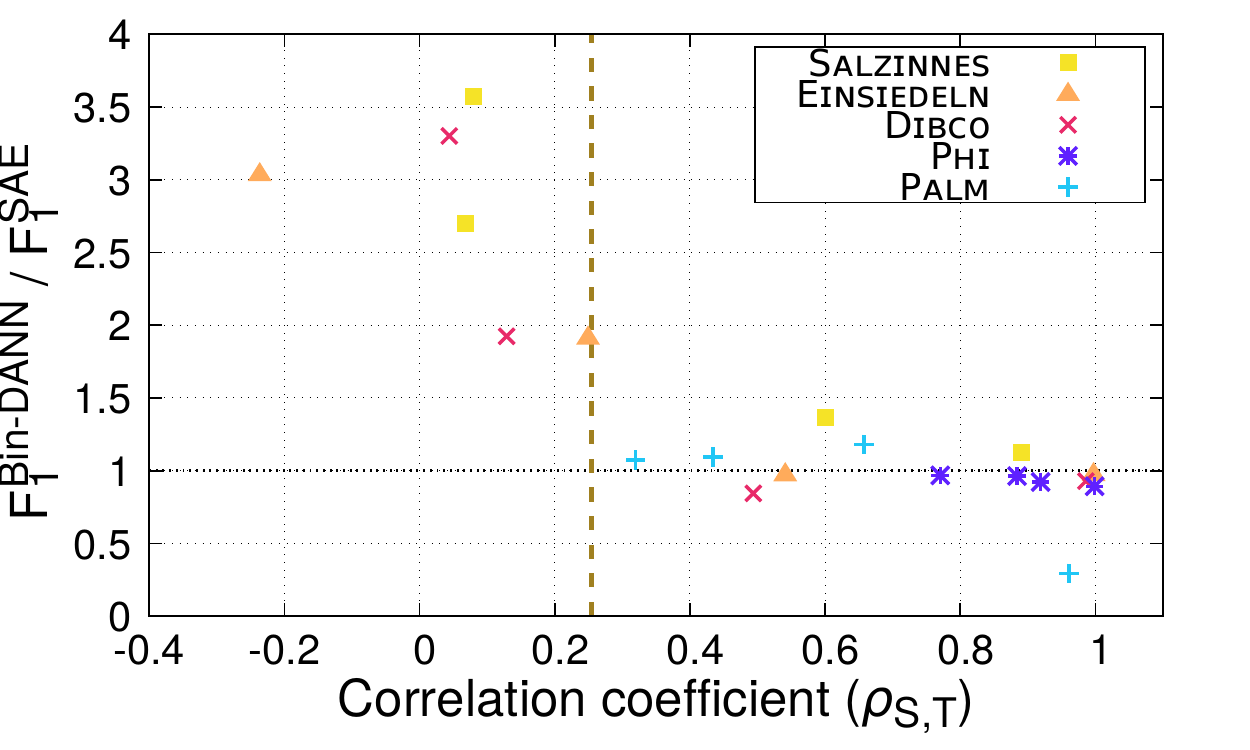}
\caption{Relationship between the correlation coefficient and the improvement obtained when applying \ac{DA}, using all datasets. The horizontal dotted line marks the boundary of no improvement with respect to the \ac{SAE} method. Higher figures are the cases in which \BinDANN{} improves the results, while the lower ones represent the cases in which it does not. The vertical dashed line symbolizes the correlation threshold applied by \AutoBinDANN{}.}
\label{fig:correl_final_factorimprovement}
\end{figure}

The relationship between the correlation \RHOst{} and the improvement obtained with \ac{DA} is shown in~\figref{fig:correl_final_factorimprovement}. Note that the selected threshold of $0.25$ splits the distribution in two parts: the first, with $\RHOth{}<=0.25$, in which the improvements obtained range from $192\%$ to $357\%$ with respect to the baseline, and the second, with $\RHOth{}>0.25$, which is composed of the lowest improvement cases and the scenarios in which the \BinDANN{} is worse than the \ac{SAE}. Note also that when $\RHOth{} \in [0.7, 0.9]$, the approach selects the \BinDANN{} model for several combinations in which it is the best choice, but also others in which the \ac{SAE} would be the best option.

\tabref{tab:average_results} includes a summary of the average of the results shown in \tabref{tab:full_results}. While the state of the art obtains $53\%$ of performance, \BinDANN{} improves it with $61.5\%$ and \AutoBinDANN{} outperforms both with $62.9\%$. To provide a reference of the best result that could be attained when the task is performed with supervised learning, the table includes the performance achieved with \ac{SAE} for all \TargetDS{}\StoT{}\TargetDS{} (see last column), obtaining an average result of $76.5$. This corresponds to the upper bound that our proposal attempts to attain. Both \BinDANN{} and \AutoBinDANN{} are able to recover part of the loss caused by the \ac{SAE} model when \SourceDS{}$\neq$\TargetDS{}. Specifically, \BinDANN{} recovers $36\%$ of the performance, while \AutoBinDANN{} achieves over $42\%$, thus getting closer to the ideal case by almost half.

\begin{table*}[!ht]
\caption{Average results for each source corpus.}
\label{tab:average_results}
\renewcommand{\arraystretch}{0.9}
\centering
\resizebox{0.99\textwidth}{!}{
\begin{tabular}{l:ccc:c}
\hline
\SourceDS{}     & F$_{1}^\text{SAE}$ (\%) & F$_{1}^\text{\BinDANN{}}$ (\%) & F$_{1}^\text{\AutoBinDANN{}}$ (\%) & F$_{1}^\text{SAE}$ (ref. \TargetDS{} \%)\\ \hline
\Salzinnes{}  & 43.2  & \textbf{69.4}  & 60.9  & 76.2 \\
\Einsiedeln{} & 53.3 & 66.5  & \textbf{67.5}  & 77.1 \\
\Dibco{}  & 54.2 & 66.3 & \textbf{71.4} & 78.1  \\
\PhiD{}    & \textbf{47.1}  & 44.2 & \textbf{47.1}  &  79.1 \\
\Palm{}   & \textbf{67.4}  & 61.2  & \textbf{67.4}  &  72.3 \\ \hline
Average  & 53.0  & 61.5  & \textbf{62.9}  & 76.5\\ \hline
\end{tabular}
}
\end{table*}

\subsection{Qualitative evaluation}

\begin{table*}[!ht]
\caption{Selected examples from \RevNote{\Dibco{}\StoT{}\PhiD{}} and \RevNote{\Dibco{}\StoT{}\Palm{}} cases. We show the binarization obtained by the state-of-the-art model (SAE), that provided by our approach (Bin-DANN), and the ground truth. \RevNote{In both cases, \AutoBinDANN{} selects the \ac{DA} approach.}}
\label{tab:image_examples}
\renewcommand{\arraystretch}{1.01}
\centering
\resizebox{1.0\textwidth}{!}{
\begin{tabularx}{\linewidth}{ c:  Y : Y : Y : Y}
& Input & \ac{SAE} & \BinDANN{} & GT \\
\hline

\begin{tabular}[c]{@{}c@{}c@{}}\Dibco{}\\ $\downarrow$\\ \PhiD{}\end{tabular} & \vspace*{4pt} \hspace*{-6pt} \includegraphics[width=0.19\textwidth]{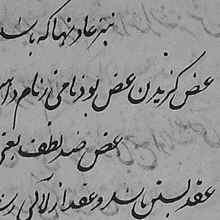} & \vspace*{4pt} \hspace*{-6pt} \includegraphics[width=0.19\textwidth]{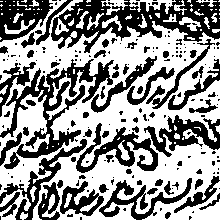} & \vspace*{4pt} \hspace*{-6pt} \includegraphics[width=0.19\textwidth]{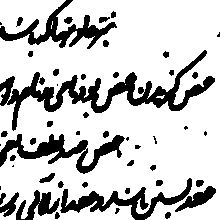} & \vspace*{4pt} \hspace*{-6pt} \includegraphics[width=0.19\textwidth]{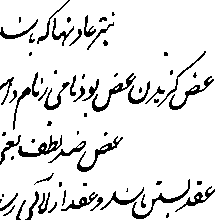}\\ \cdashline{1-5}

\begin{tabular}[c]{@{}c@{}c@{}}\Dibco{}\\ $\downarrow$\\ \Palm{}\end{tabular} & \vspace*{4pt} \hspace*{-6pt} \includegraphics[width=0.19\textwidth]{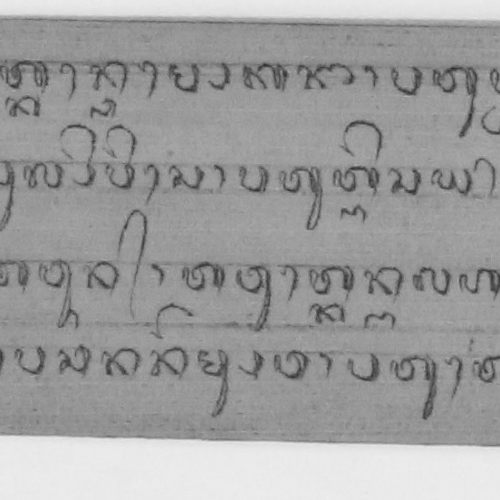} & \vspace*{4pt} \hspace*{-6pt} \includegraphics[width=0.19\textwidth]{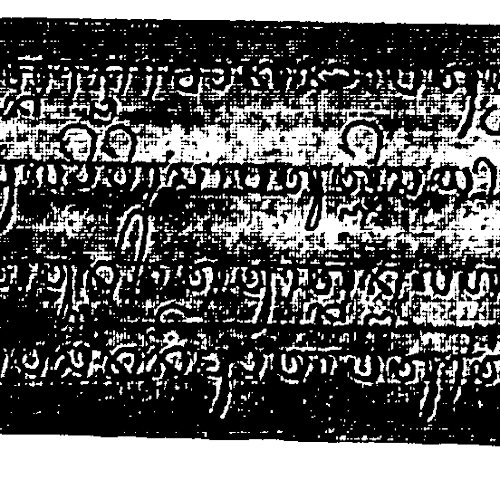} & \vspace*{4pt} \hspace*{-6pt} \includegraphics[width=0.19\textwidth]{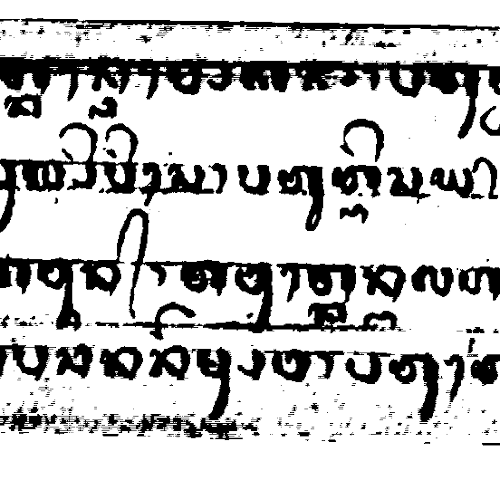} & \vspace*{4pt} \hspace*{-6pt} \includegraphics[width=0.19\textwidth]{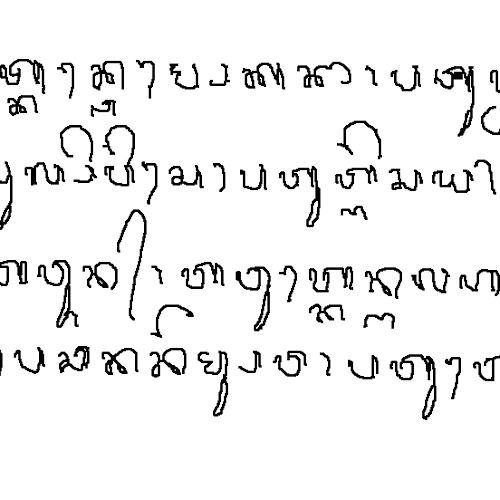}\\ 

\hline
\end{tabularx}
}
\end{table*}

To conclude our experimental section, we present some \RevNote{representative} examples of image binarization, comparing the different approaches. \tabref{tab:image_examples} shows the binarization results obtained using \RevNote{\Dibco{} as \SourceDS{} and images from \PhiD{} and \Palm{} as \TargetDS{}. In both cases, the correlation between \SourceDS{} and \TargetDS{} is low, with values of 0.13 and 0.04 for \PhiD{} and \Palm{}, respectively. Therefore, the \AutoBinDANN{} selects the \ac{DA} approach. As observed in \secref{sec:final_evaluation}, lower correlation coefficients are usually aligned to higher improvements in performance by \BinDANN{}, with the \ac{SAE} method providing poor-quality binarization.}

\RevNote{If we examine the input image of the first example, i.e. \Dibco{}\StoT{}\PhiD{}, we see that the ink bleeds through the paper from the reverse side but, given that this is noise, these pixels are labeled as background in the ground truth. However, the state-of-the-art approach is not able to differentiate this from the actual text of the page. \BinDANN{}, by adapting to the new domain, deals better with the situation, obtaining a binarized image that is much closer to the ground truth.
Something similar happens in the second example. In this case, it is observed that \Palm{} documents have a very low contrast between the ink and the background, which does not happen in \Dibco{}. This confuses the state-of-the-art method, which produces many false positives, failing to differentiate the background from the ink. Again, the proposed method manages to better deal with the issue thanks to the adaptation process.
}

\RevNote{
If we visually analyze the domains, we may observe graphic similarities and differences between them, such as in \Dibco{} and \PhiD{}, which present similar intensity in ink pixels but have differences in background color (see \figref{fig:examples_corpora}). However, the performance of binarization depends on the features learned by the network from the source domain, which do not have to coincide with our visual appreciation. For example, according to \tabref{tab:full_results}, \PhiD{}\StoT{}\Dibco{} obtains a correlation of 0.88, which means that both domains are quite similar (according to our measure); however, if we analyze the inverse case (\Dibco{}\StoT{}\PhiD{}), the correlation drops to 0.04. Therefore, the features on which the network is focusing on, can be, and certainly are, different to the features that the human eye is perceiving. 
This reinforces the use of the proposed \AutoBinDANN{} method, since it is based on the result obtained by the network when using the learned features, analyzing whether these features are also suitable to process a certain target domain. 
}

\section{Conclusions}
\label{sec:conclusions}

In this paper, we propose an unsupervised neural network approach with which to binarize images by means of adversarial training from a domain whose ground truth is not available. The approach employs a state-of-the-art method based on \acf{SAE} as its basis and makes use of a \acf{DA} artifact denominated as \acf{GRL}. This makes it possible to learn a common feature representation in order to binarize both the labeled and unlabeled domains by penalizing those features that differentiate the domain of the input image. The model is then able to learn to binarize images of different domains with respect to that used for training. 

The results suggest that this \ac{DA} approach can be employed to address the binarization issue, and that in most cases it obtains a clear improvement. However, it has been also shown that the adaptation is not always suitable, depending on the pair of corpora considered. In order to solve this problem, we propose a method that makes it possible to compare the similarity between the domains and determine whether applying the \ac{DA} process is appropriate. This process is performed by means of a comparison between the histogram obtained from the probability maps provided by the \ac{SAE} (trained with \SourceDS{}) for both domains \SourceDS{} and \TargetDS{}. 

The experiments were carried out with five different domains and, therefore, 20 possible combinations of pairs of domains. The results reveal that the decision to use \ac{DA} or \ac{SAE} is essential if a robust model for binarization is to be obtained in unsupervised scenarios, since there was a substantial increase in the average results from $53\%$ to $62.9\%$ of performance for all the study cases considered, approaching the upper bound by over $42\%$ when compared with the state of the art. 

After analyzing the results, we realized that our proposal has room for improvement. Although the decision algorithm proposed in this paper can make robust decisions, it is not, in some cases, perfect. Our future work will, therefore, involve studying other decision strategies based on machine learning.

\section*{Acknowledgment}
This work was supported by the Spanish Ministry HISPAMUS project TIN2017-86576-R, partially funded by the EU, and by the University of Alicante project GRE19-04. The first author also acknowledges support from the ``Programa I+D+i de la Generalitat Valenciana'' through grant ACIF/2019/042.

\bibliographystyle{elsarticle-num}
\bibliography{sample.bib}

\end{document}